\pdfoutput=1

\documentclass[11pt]{article}

\usepackage[table]{xcolor}

\usepackage[final]{acl}

\usepackage{times}
\usepackage{latexsym}

\usepackage[T1]{fontenc}

\usepackage[utf8]{inputenc}

\usepackage{microtype}

\usepackage{inconsolata}

\usepackage{graphicx}

\usepackage{amsmath}
\usepackage[capitalise]{cleveref}
\usepackage{subcaption}
\usepackage{tabularray}
\UseTblrLibrary{booktabs}
\usepackage{amssymb}
\usepackage{tcolorbox}
\usepackage{fancyvrb}
\usepackage{enumitem}
\usepackage{hyperref}

\newcommand{\base}[1]{\textsc{Base}\textsubscript{\,#1}}
\newcommand{\instruct}[1]{\textsc{Instruct}\textsubscript{\,#1}}
\newcommand{\corpus}[1]{Corpus\textsubscript{\,#1}}

\newcommand*\circled[1]{\raisebox{.5pt}{\textcircled{\scriptsize\sffamily \raisebox{-.4pt} {#1}}}}
\definecolor{inst-yellow}{HTML}{FFF9E5}
\definecolor{latxa-red}{HTML}{FCEAE8}
\definecolor{llama-blue}{HTML}{E5F2FF}
            
\title{Instructing Large Language Models for Low-Resource Languages:\\A Systematic Study for Basque}

\author{Oscar Sainz \quad Naiara Perez \quad Julen Etxaniz \quad Joseba Fernandez de Landa \\
\bf Itziar Aldabe \quad Iker García-Ferrero \quad Aimar Zabala \quad Ekhi Azurmendi \\ 
\bf German Rigau \quad Eneko Agirre \quad Mikel Artetxe \quad Aitor Soroa \\
HiTZ Center - Ixa, University of the Basque Country UPV/EHU \\
\texttt{\{oscar.sainz,a.soroa\}@ehu.eus} \\
}

\begin{document}
\maketitle
\begin{abstract}
Instructing language models with user intent requires large instruction datasets, which are only available for a limited set of languages. In this paper, we explore alternatives to conventional instruction adaptation pipelines in low-resource scenarios. We assume a realistic scenario for low-resource languages, where only the following are available: corpora in the target language, existing open-weight multilingual base and instructed backbone LLMs, and synthetically generated instructions sampled from the instructed backbone. We present a comprehensive set of experiments for Basque that systematically study different combinations of these components evaluated on benchmarks and human preferences from $1,680$ participants. Our conclusions show that target language corpora are essential, with synthetic instructions yielding robust models, and, most importantly, that using as backbone an instruction-tuned model outperforms using a base non-instructed model. Scaling up to Llama 3.1 Instruct 70B as backbone, our model comes near frontier models of much larger sizes for Basque, without using any Basque instructions. We release code, models, instruction datasets, and human preferences to support full reproducibility in future research on low-resource language adaptation.\footnote{\href{https://github.com/hitz-zentroa/latxa-instruct}{github.com/hitz-zentroa/latxa-instruct}}
\end{abstract}

\section{Introduction}

Large Language Models (LLMs), particularly open models, remain predominantly English-centric, with limited coverage for the vast majority of the world's languages. Despite recent efforts to incorporate additional languages during the pretraining of open LLMs, significant performance disparities still persist. Even the latest instruction-tuned models demonstrate markedly degraded capabilities when handling low-resource languages~\citep{grandury2025leaderboard}. Critically, the English-focused nature of post-training processes has widened the performance gap between languages when comparing base and instruction-tuned models.

\begin{figure}[t]
    \centering
    \includegraphics[width=\linewidth,trim={0 15pt 0 15pt},clip]{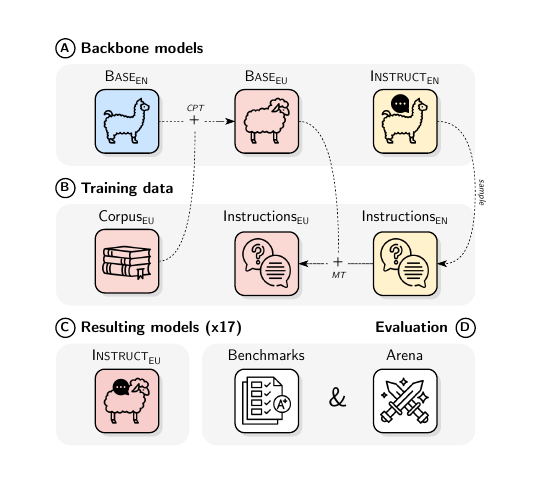}
    \caption{Systematic exploration of instruction-tuning strategies for low-resource languages. Our framework consists of: \circled{A} three backbone models (an existing base model, a continued-pretrained base model on the target language, and an existing instruct model); and \circled{B} different training data combinations, including target language corpora and synthetic instructions sampled and/or translated with the backbone models. We train \circled{C} experimental models from all possible combinations of these components, and perform \circled{D} comprehensive evaluation through both static benchmarks and human preferences to identify optimal adaptation paths.}
    \label{fig:experimental_overview}
\end{figure}

To overcome these limitations, open models can be adapted to new languages through continued training with limited resources~\cite{etxaniz-etal-2024-latxa}. In the case of instruction-tuned models in particular, various efforts have emerged that typically follow a sequential approach~\cite{ouyang2022training}: first adapting the base model through continued pretraining, then performing instruction tuning. While this multi-step process has become standard practice, little exploration has investigated alternative adaptation strategies. We question whether instruction-following capabilities could be directly transferred to new languages without dedicated instruction data, and whether instruction-tuned models could be adapted through continued pretraining, similar to base models. 

Specifically, this work systematically explores diverse strategies beyond the conventional pipeline for developing instruction-tuned models for low-resource languages, seeking to identify optimal adaptation paths for Basque as our primary case study (see~\cref{fig:experimental_overview}). We deliberately constrain our exploration to resources either readily available or creatable using open models, avoiding reliance on distillation from commercial state-of-the-art systems. While our investigation focuses on a single language, our findings likely generalize to many similarly-resourced languages: Basque represents an ideal test case, ranking approximately 50\textsuperscript{th} in Common Crawl with a presence approximately $1,000$ times smaller than English,\footnote{\href{https://commoncrawl.github.io/cc-crawl-statistics/plots/languages.html}{commoncrawl.github.io/cc-crawl-statistics/plots/languages.html}} and notably lacking pre-existing instruction datasets. This resource profile mirrors challenges faced by numerous other low-resource languages worldwide.

In addressing this research question, we further confront a key challenge in instruction-following LLM assessment: automated metrics often miss capabilities that matter to users~\cite{salakhutdinov2024chatbot}. Thus, we developed an evaluation framework combining traditional benchmarks with a crowdsourced LLM arena, where we mobilized the Basque-speaking community in a large-scale evaluation effort that gathered over $12,000$ preference annotations from $1,680$ participants. This initiative constitutes the largest human evaluation effort for a low-resource language to date.

Through this evaluation, our systematic exploration produced three key insights for developing instruction-tuned models in low-resource languages: \textbf{(1)} target language corpora is essential for performance---models lacking exposure to plain Basque text showed degradation regardless of other techniques; \textbf{(2)} while both monolingual and bilingual instruction datasets showed benefits, the latter produced consistent results across benchmark and human evaluations; and \textbf{(3)} starting from an instruction-tuned English model outperformed the approach of a base model learning to follow instructions, challenging the standard pipeline applied to low-resource languages.

In addition to these primary findings, our work makes the following contributions to the field:
\textbf{(4)} the first release of an instruction-tuned family of LLMs specifically for Basque, in sizes of 8B and 70B parameters, the latter of which proved competitive with GPT-4o and Claude Sonnet in the arena; 
\textbf{(5)} the release of large-scale, synthetic instruction-tuning datasets in English and Basque; and 
\textbf{(6)} the release of the first preference dataset in Basque, containing real user prompts, model responses, and preference annotations that could support future preference alignment research. 
Through these contributions, we aim to advance the state of language technology for Basque while establishing methodologies applicable to other low-resource languages.

\section{Related Work}

Research on developing LLMs for under-resourced languages has explored various approaches, with varying degrees of success. Initial attempts to develop models from scratch for specific low-resource languages have proven challenging due to limited training data. Multilingual model development has emerged as a more promising strategy, with researchers leveraging cross-lingual transfer learning to improve performance \cite{workshop2023bloom176bparameteropenaccessmultilingual,le-scao-etal-2022-language,shliazhko-etal-2024-mgpt}. The most effective approach to date involves continued pretraining of existing multilingual models, which allows for language-specific adaptation while benefiting from the rich linguistic representations of larger training corpora~\cite{etxaniz-etal-2024-latxa,luukkonen-etal-2023-fingpt,khanh-tungtran2024}. While progress has been made in developing these base models, optimal methods for instructing and fine-tuning them for under-resourced languages remain largely unexplored \cite{gonzalezagirre2025salamandratechnicalreport,MARTINS202553,ustun-etal-2024-aya}.

Instruction-tuning for under-resourced languages has explored various approaches to overcome the scarcity of native instruction data. Different studies leverage either English-centric pretrained models or multilingual models as pivot architectures for cross-lingual transfer \cite{purason-etal-2025-llms}. Regarding the data, researchers have explored incorporating multilingual instruction datasets that include limited coverage of lower-resourced languages \cite{shaham-etal-2024-multilingual}; translating existing English instruction sets into target languages either automatically or with human verification~\cite{joshi-etal-2025-adapting,zosa-etal-2025-got}; and applying data augmentation techniques like back-translation, language-specific prompting, and template-based instruction generation to expand limited resources \cite{li-etal-2024-x}. 
Additionally, cross-lingual in-context learning has shown interesting results \cite{cahyawijaya-etal-2024-llms}. 

Regarding Basque language adaptation, two significant studies have been conducted.~\citet{etxaniz-etal-2024-latxa} developed Latxa by adapting Llama 2 models through continued pretraining. Meanwhile, \citet{corral-etal-2025-pipeline} created Llama-eus by adapting Llama 3.1 and subsequently performing both instruction tuning and preference alignment using machine-translated data, adhering to widely accepted methodologies. However, the former focuses solely on foundation models, without considering instruction-tuned models, while the latter implements only a single strategy for instructing an adapted language model. In contrast, this work explores multiple strategies and combinations systematically to effectively instruct (or adapt) a language model for Basque.

\section{Resources}

Instructing LLMs typically relies on two components: base (or foundational) LLMs and instruction datasets. For non-hegemonic languages, obtaining instruction datasets can be very challenging, particularly in low-resource language scenarios. In the case of Basque in particular, there are no manually generated, or even good quality automatically generated, large sets of instruction-answer pairs. Consequently, as shown in \cref{fig:experimental_overview}, our available resources are constrained to corpora on the target language, and base and instruct models for high-resource languages. From these limited resources, we derive the necessary components to create Basque instruction-tuned models through strategic combinations of synthetic data generation and model adaptation. In the following sections, we describe these seed resources and derivations.

\subsection{Basque corpora} 
For the pretraining data, we have leveraged the corpora used to train Latxa, the first family of LLMs trained specifically for Basque~\cite{etxaniz-etal-2024-latxa}. This corpus comprises 4.3M of high-quality documents in Basque, roughly 3.5B Llama 3.1 tokens. Among the sources, it contains high-quality news data extracted using ad-hoc scrapers~\cite{artetxe-etal-2022-corpus}, Wikipedia\footnote{The \texttt{20231101} dump corresponding to November 2023.} and sources based on Common Crawl such as CulturaX~\cite{nguyen-etal-2024-culturax}, Colossal OSCAR~\cite{abadji-etal-2022-towards} and HLPT v1.1~\cite{de-gibert-etal-2024-new}. This corpus comes normalized, deduplicated and filtered. The data is publicly available in the HuggingFace hub.\footnote{\href{https://hf.co/datasets/HiTZ/latxa-corpus-v1.1}{hf.co/datasets/HiTZ/latxa-corpus-v1.1}} We will henceforth refer to this corpus as \corpus{EU}.

\subsection{Backbone models} 
As our base LLM (i.e., models that have not been fine-tuned to follow chat-style instructions) we use Llama 3.1~\cite{grattafiori2024llama}. Llama 3.1 is a publicly available model widely adopted by the community due to its strong performance across English and other high-resource languages. We refer to this model as \textsc{Base}\textsubscript{EN} throughout the paper. In addition, following \citet{etxaniz-etal-2024-latxa}, we train a new Latxa model based on Llama 3.1, which we denote as \textsc{Base}\textsubscript{EU}. 
For the instruction-tuned models, we adopt a similar strategy and use the instruction-following version of Llama 3.1, which we refer to as \textsc{Instruct}\textsubscript{EN}.

\subsection{Instruction Sampling and Translation}

Existing (English) instruction datasets rely on either high-quality, manually crafted instructions and responses (e.g., No Robots),\footnote{\href{https://hf.co/datasets/HuggingFaceH4/no_robots}{hf.co/datasets/HuggingFaceH4/no\_robots}} fully automatically generated instructions and responses~\cite{ding-etal-2023-enhancing, ge2024scaling}, or a combination of both, such as manually written prompts paired with automatically generated responses~\cite{zhao2024wildchat}. Using any of these datasets would introduce an additional confounding factor into our analysis (namely, knowledge distilled from a powerful LLM), which could lead a model trained on such data to outperform our \textsc{Instruct}\textsubscript{EN}, thus introducing noise into our evaluation. This would raise a separate research question that falls outside the scope of this paper: what is the best (combination of) instruction dataset(s) to train a model on? In the case of Basque, however, there is no publicly available set of instructions. The following paragraphs detail the process of generating the instructions for each language.

\paragraph{English instructions.} To avoid external influences, we instead sample instructions directly from our \textsc{Instruct}\textsubscript{EN} model. We generate the English instructions following~\cite{xu2024magpie}. Using this technique, we conditioned \textsc{Instruct}\textsubscript{EN} to generate instructions of different types and tasks: \textit{general-purpose}, \textit{code}, \textit{math}, \textit{arithmetic} and \textit{translation}. We generated a total of 4M English instructions. However, after a hyperparameter search, we found out that using just 1M instructions yielded better results overall (see~\cref{ap:training_details}). We share more details and examples of the process in~\cref{ap:instruction_generation}.

\paragraph{Basque instructions.} \label{par:basque_inst} 
We translated instructions sampled from \textsc{Instruct}\textsubscript{EN} using few-shot prompting with \base{EU}. Existing machine translation systems for the English--Basque language pair (e.g., NLLB~\cite{costa2022no}) are primarily trained on sentence-level textual data and often struggle with more complex inputs, including selectively translating natural language content embedded within code snippets. By leveraging an LLM like \base{EU}, which has been exposed to diverse data types, we obtained higher-quality translations for this setting. Moreover, using a model trained within our own experimental framework allows us to avoid introducing external factors into our pipeline. More details about the process and prompts used to translate the instructions are given in~\cref{ap:instruction_generation}.

\section{Experimental Setup} 
\label{ssec:training_configurations}

We formalize our experimental setup as follows. Let $\mathcal{M} = \{\text{\base{EN}}, \text{\base{EU}}, \text{\instruct{EN}}\}$ be the set of backbone models and $\mathcal{D} = \{\text{Corpus\textsubscript{\,EU}}, \text{Instructions\textsubscript{\,EN}}, \text{Instructions\textsubscript{\,EU}}\}$ be the set of binary variables indicating whether to use Basque corpora, English instructions, and/or Basque instructions. The space of possible configurations is thus $\mathcal{M} \times \mathcal{P}(\mathcal{D})$, where $\mathcal{P}(\mathcal{D})$ is the power set of $\mathcal{D}$, yielding $|\mathcal{M}| \times 2^{|\mathcal{D}|} = 3 \times 2^3 = 24$ theoretical combinations. Note that we explore training strategies that leverage both \textbf{raw text and instruction data simultaneously}. From the total of 24 combinations, we exclude redundant configurations where a model is retrained on data it was originally trained with. The resulting set of distinct instruction-tuned model variants therefore comprises 18 configurations: the original Llama 3.1 Instruct 8B (i.e., \instruct{EN}) and 17 new 8B-sized models.~\cref{tab:model-variants} in~\cref{ap:training_details} provides a complete account of all model variants and their shorthand names. Additionally, we trained a 70B model following the configuration that performed best in preliminary benchmark evaluations.

Regarding the baselines, the primary baseline in our analysis is the \instruct{EN} model, as it is the only backbone capable of following instructions. However, since we examine the effect of each variable in $\mathcal{D}$ individually, the specific points of comparison used vary across cases. For additional context, we also evaluate two proprietary models known for their strong performance in Basque:\footnote{We did not include the instructed Llama-eus~\cite{corral-etal-2025-pipeline} in our evaluation, as it was not publicly available at the time of experimentation.} OpenAI’s GPT-4o\footnote{\texttt{gpt-4o-2024-11-20}} and Anthropic’s Claude 3.5 Sonnet.\footnote{\texttt{claude-3-5-sonnet-20241022}} 

\section{Evaluation}

We employed two complementary evaluation approaches to assess the impact of each instruction-tuning recipe. On the one hand, we used a selection of static benchmarks that evaluate specific model capabilities and knowledge through standardized tests. On the other hand, we conducted human evaluations through A/B testing (arena style) to capture qualitative aspects of model performance. In addition, we look into the impact of our recipes on safety and bias.

\subsection{Static Benchmarks}
\label{ssec:static_benchmarks}

We selected benchmarks that are close to real use cases, from a varied range of categories: \textit{reading comprehension}, \textit{common sense}, \textit{linguistic proficiency}, \textit{knowledge} and \textit{maths \& reasoning}.

For each benchmark, where possible, we evaluated the Basque, English, and Spanish versions to facilitate the analysis of language-specific tradeoffs for each fine-tuned model variant. This choice of evaluation languages reflects the linguistic reality of the Basque-speaking community in northern Spain, where Basque and Spanish are co-official and English is the most commonly taught foreign language. Importantly, these languages come from distinct families: Basque is a language isolate, Spanish is Romance, and English is Germanic. Thus, we examine cross-lingual transfer effects and assess whether improvements in our language come at the cost of performance in related community languages, including one---Spanish---not directly targeted by our experiments. 
In total, then, we evaluated 27 benchmarks, as detailed in~\cref{sap:static-benchmarks}.

For conducting these evaluations, we relied on LM Evaluation Harness~\cite{biderman2024lessons}. Most datasets are framed as multiple-choice problems where models' answers are determined by selecting the option with the highest log probability. For generative tasks, answers are directly sampled from the model. To provide models with contextual examples, our evaluations employed a few-shot setting. All results are measured for accuracy following standard, public implementations. Refer to~\cref{sap:static-benchmarks} and our repository for details.

When evaluating proprietary models, we cannot directly compute log probabilities because we have no access to model weights. This limitation restricts our evaluation to only those benchmarks implemented as explicit letter-choice questions (A, B, C, ...) and the free-form generative task MGSM, excluding benchmarks that require comparing verbalized option likelihoods. For the compatible multiple-choice benchmarks, we prompt models to output a single letter as their answer, using the same prompts and few-shot examples as with open models to maintain comparability.

\subsection{Human Evaluation: Arena}
\label{ssec:arena}

Unlike static benchmarks, which rely on fixed datasets and automatic metrics, arena-style evaluations are better suited for assessing open-ended text generation, where subjective quality judgments play a central role. In this section, we first describe our implementation of the arena framework, including details on participants and evaluation conditions. For additional details about the human evaluation, refer to~\cref{sap:arena-details}, where we describe our infrastructure and introduce the Bradley--Terry model~\cite{bradly-terry}, which we use to infer a model ranking from the collected pairwise preferences.

\begin{figure}[t]
    \centering
    \includegraphics[width=.5\textwidth,trim={8pt 0 0 0},clip]{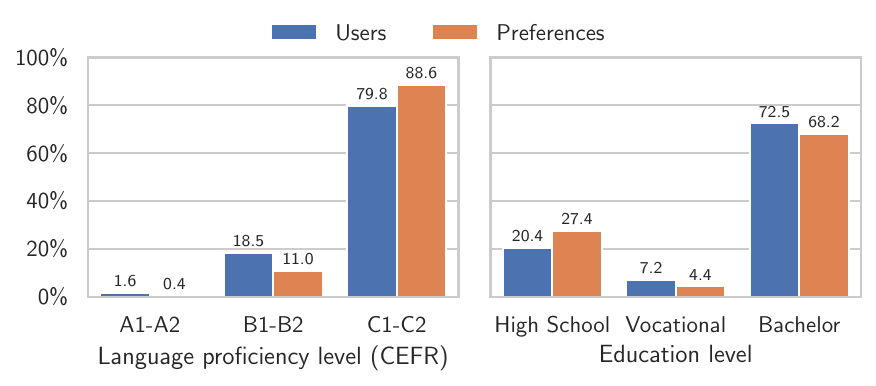}
    \caption{Distribution of participants and preferences by education level and language proficiency level.}
    \label{fig:arena-participants}
\end{figure}

To gather human preferences for our evaluation, we organized a community-driven initiative. This collaborative effort ran for 14 days and attracted approximately 1,680 participants, resulting in a total of 12,890 preference annotations. The event was open to any Basque speaker, regardless of their proficiency level. Participants were required to register their educational background and language proficiency before contributing. Once registered, users could submit prompts and compare model responses.~\cref{fig:arena-participants} shows that the majority of the participants---and, therefore, the preferences---have a bachelor or superior education and a high or native language proficiency level.

In the annotation process, participants evaluated pairs of model responses by making a three-way choice (i.e., prefer model `A', prefer model `B', or consider them tied) across two dimensions: content quality and linguistic quality. Linguistic quality was considered as a separate dimension because not all models produce fluent and sound Basque---an uncommon issue among high-resource languages. In cases where participants' judgments were contradictory between the dimensions, a third question about overall quality was presented to determine the final choice.

It merits mention that we offered prizes based on user activity to encourage widespread participation and maximize the number and diversity of collected preferences. While this strategy succeeded in increasing engagement, it also attracted malicious or dishonest users who prioritized quantity over quality to win rewards. We proactively identified and banned such actors using a combination of heuristics and manual review.

\begin{figure*}[t]
    \begin{subfigure}[b]{0.5\textwidth}
        \includegraphics[width=\textwidth]{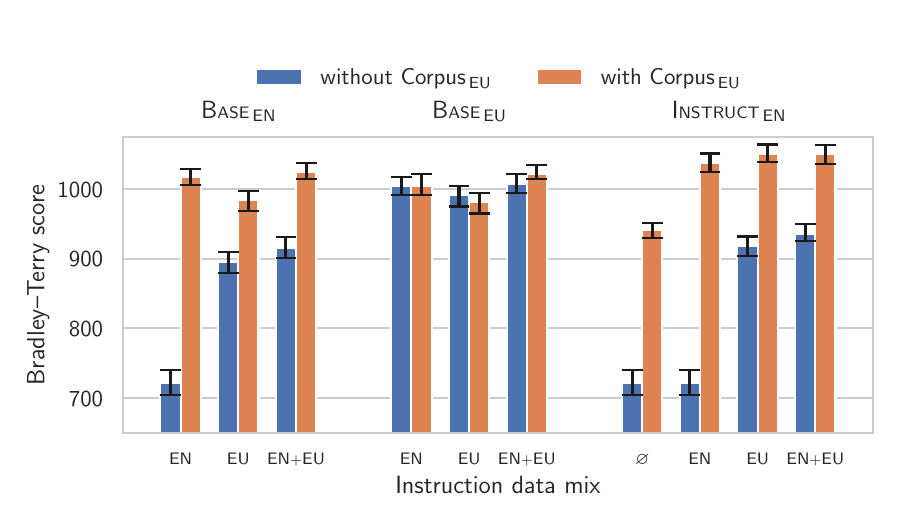}
        \caption{Bradley--Terry scores in the human evaluation arena}
        \label{fig:main_comparison_data-ELO-bars}
    \end{subfigure}%
    \begin{subfigure}[b]{0.5\textwidth} 
        \includegraphics[width=\textwidth]{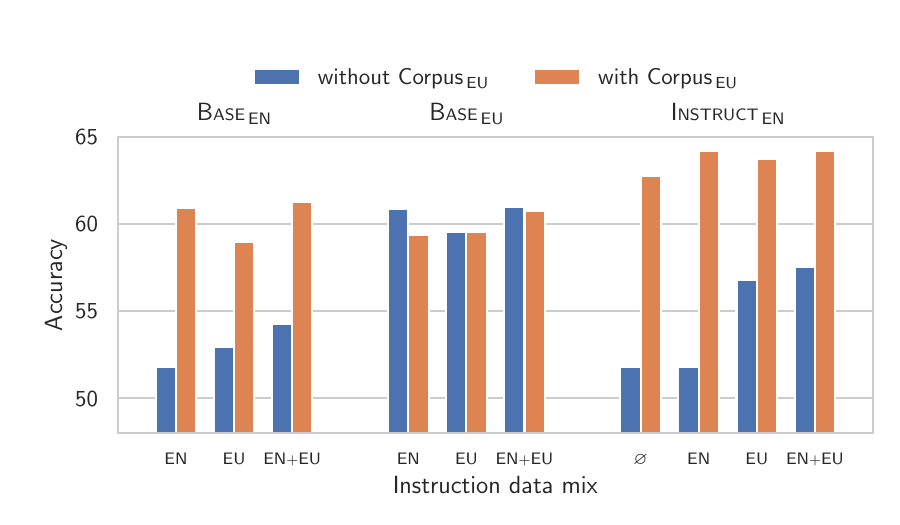}
        \caption{Average accuracy results in Basque benchmarks}
        \label{fig:main_comparison_data-acc-bars}
    \end{subfigure}%
    \caption{Performance comparison of instruction-tuned models across different dimensions: backbone model, Corpus\textsubscript{\,EU} usage and instruction data composition. Error bars in~\cref{fig:main_comparison_data-ELO-bars} indicate 90\% confidence intervals.}
    \label{fig:main_comparison}
\end{figure*}

\subsection{Safety and Bias}
\label{ssec:safety-and-bias}

This paper does not address the alignment step typically required to reduce biases and prevent unsafe responses in production-ready LLMs. However, the backbone model \instruct{EN} is already aligned, and we expect variants based on this backbone to retain safety-related behaviors. To verify this, we constructed a new Basque-English parallel dataset inspired by XSTest~\cite{rottger-etal-2024-xstest}, combining clearly unsafe prompts with superficially similar but safe ones. We measure both Violation rates (VR) and False Refusal rates (FRR), where the model wrongly declines safe prompts. For bias, we rely on BBQ~\cite{parrish-etal-2022-bbq} and its adaptation to Basque~\cite{saralegi-zulaika-2025-basqbbq}, reporting results in terms of accuracy. Further details are available in~\cref{ap:safety-bias}.

\section{Results}
\label{sec:results}

In the following paragraphs, we disclose the effect of each component: backbone models, Basque corpora and bilingual instructions. For both the benchmark and the human evaluation, main results are presented in~\cref{fig:main_comparison}. Complete results for all tested models, broken down by benchmark and language, along with detailed arena evaluation results, can be consulted in~\cref{ap:complete-results}.

\subsection{The Impact of Basque Corpora}  
\label{ssec:impact-corpus}

We begin by analyzing the influence of \corpus{EU}, as it is intuitively the most critical resource for teaching a new language to an LLM.~\cref{fig:main_comparison} confirms that this intuition aligns well with empirical results. In both human evaluations (\cref{fig:main_comparison_data-ELO-bars}) and benchmark scores (\cref{fig:main_comparison_data-acc-bars}), models trained on \corpus{EU} achieved significantly better performance. The advantage is especially pronounced---up to 12 points in accuracy and over 300 points in arena score---when no other Basque signal (i.e., Basque instructions) is included. However, for models that already use \base{EU} as their backbone, additional exposure to \corpus{EU} offered little benefit and was sometimes even detrimental.

We conclude that \textbf{using target language corpora is highly beneficial and possibly essential for training an instruction-tuned LLM in our low-resource language}. Therefore, the following analyses will focus exclusively on the variants trained with \corpus{EU}.

\begin{table*}[t]
    \centering
        \resizebox{\textwidth}{!}{%
\begin{tabular}{lcccccccc}
\toprule
                & \multicolumn{4}{c}{\textbf{8B}} & \multicolumn{2}{c}{\textbf{70B}} & \multicolumn{2}{c}{\textbf{Proprietary}} \\
                \cmidrule(lr){2-5} \cmidrule(lr){6-7} \cmidrule(lr){8-9}
       & \textbf{\instruct{EN}} & \textbf{+ C\textsubscript{\,EU}\,I\textsubscript{\,EN}} & \textbf{+ C\textsubscript{\,EU}\,I\textsubscript{\,EU}} & \textbf{+ C\textsubscript{\,EU}\,I\textsubscript{\,EN+EU}} & \textbf{\instruct{EN}} & \textbf{+ C\textsubscript{\,EU}\,I\textsubscript{\,EN}} & \textbf{3.5 Sonnet} & \textbf{GPT-4o} \\
\midrule

Belebele                       & 73.89 & 80.00 & 81.44 & \textbf{83.00} & 89.11 & \textbf{91.00}  & \textbf{94.22} & 92.88 \\
BertaQA\textsubscript{\,Global}  & 67.10 & \textbf{74.62} & 73.54 & 72.99 & 83.53 & \textbf{87.42} & \textbf{93.52} & 91.01 \\
BertaQA\textsubscript{\,Local}   & 44.97 & 65.23 & \textbf{66.07} & 65.57 & 53.51 & \textbf{77.71} & \textbf{80.45} & 74.83 \\
EusProficiency                 & 34.13 & \textbf{52.83} & 52.06 & 52.35 & 43.59 & \textbf{68.00} & \textbf{81.60} & 74.25 \\
EusReading                     & 49.72 & 59.66 & \textbf{62.78} & 61.93 & 72.16 & \textbf{78.98} &  \textbf{87.39}     & 84.38 \\
EusTrivia                      & 45.01 & 61.05 & \textbf{62.33} & 62.10 & 62.51 & \textbf{74.17} & \textbf{84.60} & 80.70 \\
EusExams                       & 46.21 & 56.00 & 56.01 & \textbf{56.23} & 63.28 & \textbf{71.56} & \textbf{82.68} & 79.17 \\
MGSM                           & 45.60 & \textbf{54.00} & 46.40 & 50.80 & 76.40 & \textbf{80.00} & \textbf{85.20} & 79.20 \\
MMLU                           & 50.37 & \textbf{57.04} & 52.96 & 56.30 & 68.52 & \textbf{68.89} & \textbf{79.63} & 76.66 \\
\rowcolor{gray!10} \textbf{Benchmark Avg} & 50.78 & 62.27 & 61.51 & \textbf{62.36} & 68.07 & \textbf{77.53} & \textbf{85.48} & 81.45 \\

\midrule
Arena\textsubscript{\,Content} & 766 {\small (-17,+14)} &  1031 {\small (-12,+15)} & 1045 {\small (-13,+11)} & 1047 {\small (-12,+12)} & - & 1127 {\small (-11,+10)} & 1150 {\small (-17,+12)} & 1183  {\small (-13,+15)} \\
Arena\textsubscript{\,Language} & 783 {\small (-12,+12)} & 1036 {\small (-10,+11)} & 1034 {\small (-10, +8)} & 1038 {\small ( -8,+10)} & - & 1083 {\small (-13,+13)} & 1082 {\small (-11,+11)} & 1093 {\small (-10,+12)} \\
\rowcolor{gray!10} \textbf{Arena\textsubscript{\,Global}} & 722 {\small (-17,+19)} & 1038 {\small (-13,+13)}  & 1050 {\small (-11,+14)} & 1050 {\small (-14,+13)} & - & 1141 {\small (-11,+15)} & 1153 {\small (-21,+13)} & 1188 {\small (-17,+13)} \\
\bottomrule
\end{tabular}
}

    \caption{Results for the baseline (\instruct{EN}), the best 3 performing variants, 70B models and proprietary models. Best results among comparable setups are marked in \textbf{bold}. Arena scores are given with 90\% confidence intervals.}
    
    \label{tab:results}
\end{table*}

\subsection{The Impact of Instruction Data}  
\label{ssec:impact-instructions}

We analyze the effect of instruction data by comparing variants trained with no instructions ($\varnothing$), English-only instructions (I\textsubscript{\,EU}), Basque-only instructions (I\textsubscript{\,EN}), and their combination (I\textsubscript{\,EN+EU}).

Starting with the question of whether to use instructions at all, we focus on \instruct{EN}-based variants. Human evaluation results (\cref{fig:main_comparison_data-ELO-bars}) clearly show that incorporating instructions, regardless of language, helps mitigate catastrophic forgetting and improves arena scores by nearly 100 points. In contrast, benchmark results (\cref{fig:main_comparison_data-acc-bars}) show only marginal gains, particularly when using Basque instructions. This discrepancy highlights the limitations of static benchmarks and underscores the value of human or text generation-based evaluation.

When comparing instruction languages, we observe a general trend: English instructions tend to yield better results across both evaluation methods. However, there are exceptions. For example, models based on \instruct{EN} perform comparably or slightly better with Basque instructions in human evaluations, while \base{EU}-based models perform similarly on benchmarks. 

Notably, combining English and Basque instructions consistently produces the best results across most scenarios. While this improvement could be attributed to certain model variants having access to more training data, our preliminary results in~\cref{ap:training_details} refute this hypothesis, as using more monolingual instructions (1M vs 4M) resulted in similar results.

Although the improvements are not always significantly better, we conclude that \textbf{including instructions in both languages results in more robust models}, achieving stronger performance regardless of the backbone. Consequently, the remainder of our analysis will focus on models trained with bilingual instruction data.

\subsection{The Impact of Backbone Models and Curriculum Learning}
\label{ssec:impact-curriculum}

By analyzing models trained from different backbones, we explore various curriculum learning strategies: (i) teaching the language first and then instruction following, (ii) teaching instruction following in English first and then the target language, or (iii) learning everything simultaneously.

\paragraph{Language first vs. simultaneously.} 
When comparing models based on \base{EN} (i.e., acquiring Basque and instruction-following capabilities simultaneously) with those based on \base{EU} (learning the language first, then instruction following), we observe no significant difference in performance. Interestingly, the \base{EU} variant \emph{without} access to \corpus{EU} during instruction tuning achieves performance nearly identical to the \base{EN} variant \emph{with} access to \corpus{EU}, across all instruction settings. This suggests that teaching the language in a separate pretraining step offers no measurable advantage. From this, we conclude that \textbf{there is no compelling reason to separate language acquisition from instruction tuning}.

\paragraph{Instructions first vs. simultaneously.}  
While instruction tuning pipelines are often complex and multi-staged~\cite{lambert2025tulu3pushingfrontiers}, our approach adopts a simpler structure. Previous work by~\citet{xu2024magpie} showed that models initialized from \base{EN} and trained with sampled instructions from \instruct{EN} perform comparably to \instruct{EN}. However, our findings indicate that this strategy does \emph{not} transfer well to low-resource languages. As shown in~\cref{fig:main_comparison}, models based on \instruct{EN} consistently outperform those based on \base{EN}, both in human evaluations and benchmarks. These results support the conclusion that \textbf{starting from a well-instructed English backbone yields better performance than learning everything from scratch}.

\subsection{The Impact of the Scale}
\label{ssec:impact-scale}

Based on our previous analyses, we scaled up the variant that performed best in preliminary benchmark evaluations.\footnote{We did this before running the arena as we wanted to include a 70B model in the human evaluation.}~\cref{tab:results} shows the results for some multiple-choice benchmarks and arena scores. On the one hand, we have the three best-performing 8B variants and the baseline. On the other hand, we present the results for the 70B best-performing variant and the baseline. Finally, we also show the performance of Claude 3.5 Sonnet and GPT-4o.

\paragraph{Scaling to bigger sizes.}
We analyzed the effect of our language adaptation process when training a larger model. Despite the results of the 70B \instruct{EN} baseline being significantly better than the 8B counterpart (even surpassing the \instruct{EU} variants), we observe that our language adaptation step still had similar improvements to those obtained with 8B sized models---almost 10 accuracy points gain on average. The biggest gains are observed in local knowledge and language proficiency, the only benchmarks where the 70B \instruct{EN} underperforms the best 8B variant.

\paragraph{Comparing to the State of the Art.}
When compared to the leading commercial models in Basque, our best model falls slightly behind in most benchmarks except for BertaQA\textsubscript{\,Local} and MGSM, where our model performs better than GPT-4o---particularly in local knowledge about the Basque Country. Regarding the arena score, \textbf{our best model is on par with the commercial models in perceived linguistic quality}, but falls behind the best model in the content quality and global scores. Interestingly, Claude 3.5 Sonnet outperforms GPT-4o in all benchmarks, but the latter gets a higher score in the arena. Our best model being worse than SotA despite focusing on Basque might be related to the weaker backbone model we used, as \instruct{EN} is overall a weaker model. Using a stronger and larger backbone model in the future could lead to results that match the SotA models in benchmarks and arena score.

\section{Analysis and Discussion}
\label{sec:analysis}

In this section, we provide additional analysis and discussion of our results. First, we focus on the correlation of our two main evaluation methods. Then, we measure the trade-off between Basque and other languages. Finally, we analyze the safety and biases of our models.

\begin{figure}[t]
    \centering
    \includegraphics[width=\linewidth]{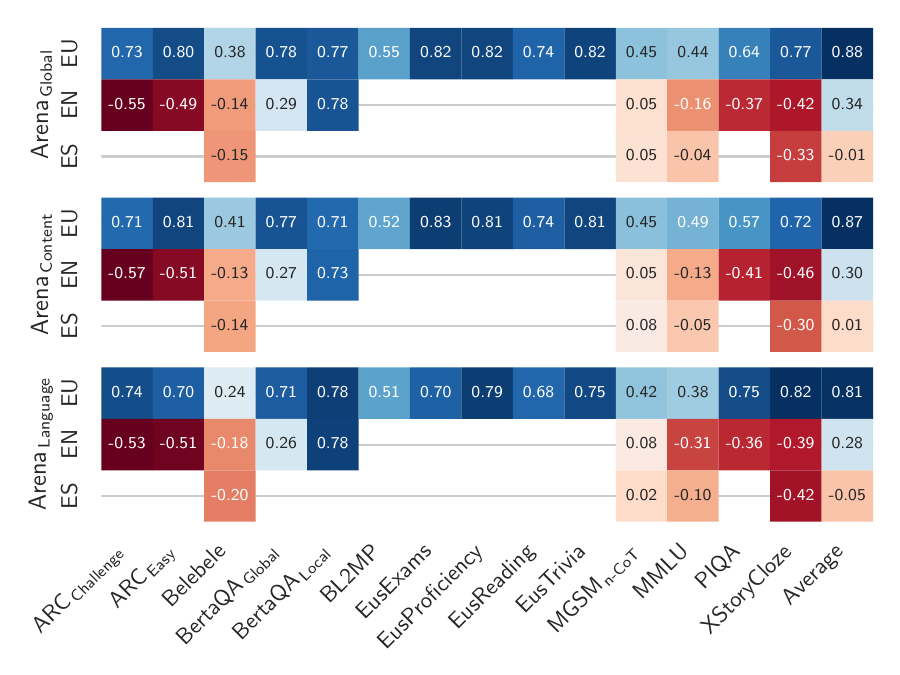}
    \caption{Spearman's rank correlation coefficients between benchmark performance and arena evaluation dimensions.}
    \label{fig:arena-benchmarks-correlation}
\end{figure}

\paragraph{Benchmark--Arena correlation.} \cref{fig:arena-benchmarks-correlation} shows Spearman's rank correlation coefficients, $\rho$, between benchmark performance and arena scores, across different benchmark languages (Basque, English, Spanish) and arena dimensions (content, language, and global). We observe that average Basque benchmark performance and arena rankings correlate strongly, with $\rho > 0.80$ across all arena dimensions---which suggests that automated Basque benchmarks may provide a reliable proxy for human evaluation in future research. This correlation is particularly pronounced for specific benchmarks, including EusProficiency, EusTrivia, EusExams and BertaQA, which are interestingly the datasets that were natively constructed in Basque, rather than translated from existing English benchmarks. The average of English benchmarks displays overall positive but non-significant correlations. Only BertaQA shows positive correlations, with the local subset obtaining correlations similar to the Basque BertaQA, likely reflecting the types of culturally-specific questions that users posed in the arena. Spanish benchmarks show, on average, no correlation with the arena. Some of the English and Spanish benchmarks show large negative correlations, reflecting the performance trade-off between Basque and other languages.

\paragraph{Trade-off between Basque and other languages.}

\begin{figure}[t]
    \centering
    \includegraphics[width=\linewidth]{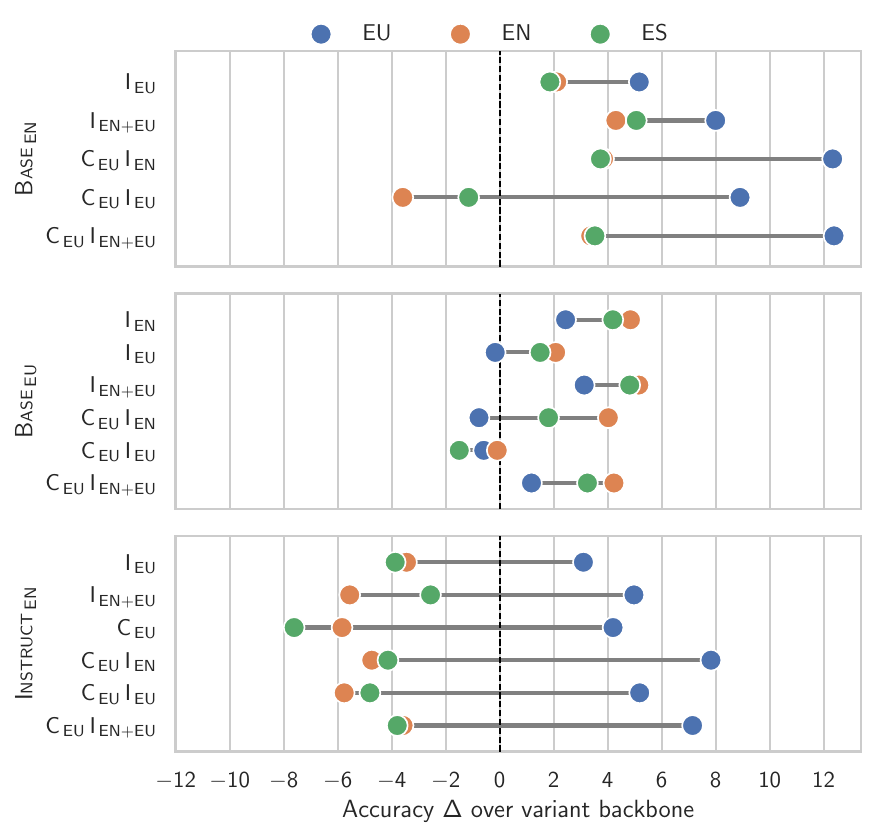}
    \caption{Accuracy differences for model variants compared to their respective backbones across Belebele, MGSM, MMLU, and XStoryCloze averages by language. Positive values indicate performance improvements; negative values indicate regression.}
    \label{fig:language-tradeoffs.dumbbell}
\end{figure}

\cref{fig:language-tradeoffs.dumbbell} shows performance changes across languages relative to each backbone model on the multilingual benchmarks Belebele, MGSM, MMLU, and XStoryCloze. We observe that \instruct{EN}-based models exhibit a clear trade-off: improvements in Basque come at a cost of decreased performance in English and Spanish, suggesting a competitive relationship between languages in the model's parameter space. In contrast, greater flexibility for multilingual adaptation is observed in \base{EN} models, which improve across all three languages (though from a lower absolute performance baseline). \base{EU} models show moderate changes with gains primarily in Spanish and English rather than Basque. As observed previously, Corpus\textsubscript{\,EU} consistently yields the largest performance gains for the target language. Among \instruct{EN} variants, the configuration with Corpus\textsubscript{\,EU} and Instructions\textsubscript{\,EN} achieves the most Basque improvement and the least regression in other languages. Despite these adaptation strategies, models still perform better in English and Spanish on equivalent tasks, with the exception of culturally-specific knowledge as evidenced by the results on the BertaQA dataset (see complete results in~\cref{sap:benchmark-results}). 

\paragraph{Safety and bias.} We evaluate the model variant INS\textsubscript{\,EN}\,C\textsubscript{\,EU}\,I\textsubscript{\,EN}, comparing it with two critical counterparts: (1) BAS\textsubscript{\,EN}\,C\textsubscript{\,EU}\,I\textsubscript{\,EN}, to analyze the impact of starting from an already instruction-tuned backbone versus a base model (8B only); and (2) \instruct{EN}, to measure potential alignment changes introduced with our fine-tuning data mix (8B and 70B).
As shown in~\cref{tab:benchmark-safety-bias-small}, the BAS\textsubscript{\,EN}\,C\textsubscript{\,EU}\,I\textsubscript{\,EN} model demonstrates low safety and high bias in the Basque language compared to English, reflecting a significant lack of alignment. 
In contrast, \instruct{EN} and its variant INS\textsubscript{\,EN}\,C\textsubscript{\,EU}\,I\textsubscript{\,EN} are better aligned in both languages. Interestingly, in Basque, the 8B \instruct{EN} shows a lower VR than both BAS\textsubscript{\,EN}\,C\textsubscript{\,EU}\,I\textsubscript{\,EN} and INS\textsubscript{\,EN}\,C\textsubscript{\,EU}\,I\textsubscript{\,EN} models and even the larger 70B counterpart. This behavior is due to the model's limited comprehension of instructions in the Basque language, leading it to produce inconsistent yet safe responses. On the other hand, the 70B models display comparable performance across both languages, with the INS\textsubscript{\,EN}\,C\textsubscript{\,EU}\,I\textsubscript{\,EN} version slightly outperforming others and achieving a lower VR in Basque. Notably, the FRR remains very low across all models, indicating that safety mechanisms do not come at the cost of excessive conservatism. Bias outcomes in English are consistently better than in Basque, and larger models generally perform better than smaller ones. However, the differences in bias between the original and language-adapted models are minimal. 
Overall, we demonstrate that much of the safety and bias alignment is transferred from \instruct{EN}, not only in the newly added language, but also in the predominant language.

\begin{table}[t]
    \centering

\begin{tblr}{
    width=\linewidth,
    colspec={ll*{6}{X[r]}},
    cells={font=\scriptsize},
    row{1,2,3}={c,font=\scriptsize\bfseries},
    column{1,2}={font=\scriptsize\bfseries},
    column{1}={rightsep=0pt},
    rowsep=0pt
}
    \toprule
    & & \SetCell[c=3]{c} EU &&& \SetCell[c=3]{c} EN \\
    \cmidrule[r=-0.4]{3-5} \cmidrule[l=-0.4]{6-8}
    & & VR & FRR & BBQ & VR & FRR & BBQ \\
    & & ↓ & ↓ & ↑ & ↓ & ↓ & ↑ \\
    \midrule
    \SetCell[r=3]{m} 8B
    & INS\textsubscript{\,EN}\,C\textsubscript{\,EU}\,I\textsubscript{\,EN} 
        & 24.00 & 0.00 & 70.80 & 16.00 & 0.00 & 87.06 \\
    & BAS\textsubscript{\,EN}\,C\textsubscript{\,EU}\,I\textsubscript{\,EN} 
        & 44.00 & 0.00 & 51.73 & 20.00 & 4.00 & 72.35 \\
    & \instruct{EN} 
        &  8.00 & 0.00 & 71.03 &  4.00 & 0.00 & 87.65 \\
    \midrule
    \SetCell[r=2]{m} 70B
    & INS\textsubscript{\,EN}\,C\textsubscript{\,EU}\,I\textsubscript{\,EN} 
        &  4.00 & 0.00  & 85.29 & 4.00 & 0.00 & 94.38 \\
    & \instruct{EN}
        & 20.00 & 0.00  & 84.90 & 8.00 & 0.00 & 95.32 \\
    \bottomrule
\end{tblr}
    \caption{Safety and bias results for Basque and English datasets. Safety is measured in terms of violation rate (VR) and false refusal rate (FRR) where ↓ indicates lower values are better. Bias is measured with BBQ accuracy, where ↑ indicates higher values are better.}
    \label{tab:benchmark-safety-bias-small}
\end{table}

\section{Conclusions}
\label{sec:conclusions}

This systematic study on instruction-tuning LLMs for Basque reveals several key strategies for low-resource language adaptation. We found that \textbf{target language corpora are essential} for effective learning. Employing \textbf{bilingual (English and Basque) synthetic instructions yielded the most robust models} whereas English-only instructions remain competitive. Crucially, \textbf{starting from an instruction-tuned English model and adapting it to Basque proved more effective} than training a base model for both language and instruction-following, or pretraining for language separately before instruction tuning. Our work contributes new Basque models, open instruction and human preference datasets, and methodological insights to guide future low-resource LLM development.

In the future, we plan to extend the exploration using instructions created by humans. The preference data we release can also be used to align the models. Scaling to stronger backbones will also lead to better results that could match the SotA commercial models in Basque. The strong correlation observed between aggregated Basque benchmarks and human evaluations also suggests a path for more efficient proxy evaluations.

\section*{Limitations}

This study presents a systematic analysis aimed at identifying the most effective method for developing an instruction-tuned model for a low-resource language. However, due to the combinatorial nature of such analyses, we had to constrain certain dimensions of our exploration, as adding any additional axis would effectively double the amount of work required, including costly human evaluations.

Our first limitation is the choice of language. We focused on Basque, primarily because it is a low-resource language with just enough available data to enable language adaptation of base models~\cite{etxaniz-etal-2024-latxa}. While some other languages have significantly fewer resources than Basque, our conclusions may not fully generalize to those more extreme low-resource scenarios.

The second limitation is the choice of model family. We conducted all experiments using the Llama 3.1 family as the backbone. This decision was motivated by its widespread adoption and its existing ability to generate text in Basque, although often with substantial linguistic errors. Evaluating more recent or higher-performing models could slightly influence our findings. However, to the best of our knowledge, there is currently no open model family capable of producing linguistically correct Basque.

Third, this study primarily focuses on the initial instruction-tuning phase. While we did collect preference data, we did not extend our analysis to include preference alignment techniques. Including this additional phase would again have doubled the experimental workload and human evaluation requirements.

Finally, although we did not perform analyses on potential data contamination issues, previous work on which our work is based took measures against contamination~\citep{etxaniz2024bertaqa, etxaniz-etal-2024-latxa}. 

\section*{Acknowledgements}

This work has been partially supported by the Basque Government (Research group funding IT1570-22 and IKER-GAITU project), the Spanish Ministry for Digital Transformation and of Civil Service, and the EU-funded NextGenerationEU Recovery, Transformation and Resilience Plan (ILENIA project, 2022/TL22/00215335; and ALIA project). We acknowledge EuroHPC JU for awarding the project ID EHPC-EXT-2024E01-042 access to CINECA Leonardo and project ID EHPC-AI-2024A04-074 access to BSC MareNostrum 5. Julen Etxaniz and Ekhi Azurmendi hold a PhD grant from the Basque Government (PRE\_2024\_2\_0028 and PRE\_2024\_1\_0035, respectively).

\bibliography{anthology,custom}

\appendix


\begin{figure*}[!ht]
    \centering
    \begin{subfigure}[b]{\textwidth}

        \begin{tcolorbox}[left=.5em]
\begin{Verbatim}[fontsize=\small,commandchars=\\\{\}]
\textbf{<|begin_of_text|>}
\textbf{<|start_header_id|>}system\textbf{<|end_header_id|>}
Cutting Knowledge Date: December 2023
Today Date: 26 Jul 2024
\textbf{<|eot_id|>}
\textbf{<|start_header_id|>}user\textbf{<|end_header_id|>}
\end{Verbatim}
        \end{tcolorbox}
        \vspace{-.5em}
        \caption{General}
        \label{fig:general_inst_prompt}
    \end{subfigure}
    \hfill
    \begin{subfigure}[b]{0.49\textwidth}
        \vspace{.8em}
        \begin{tcolorbox}[left=.5em]
\begin{Verbatim}[fontsize=\small,commandchars=\\\{\}]
\textbf{<|begin_of_text|>}
\textbf{<|start_header_id|>}system\textbf{<|end_header_id|>}
You are an AI assistant designed to provide 
helpful, step-by-step guidance on solving 
math problems. The user will ask you a wide 
range of complex mathematical questions. 
Your purpose is to assist users in 
understanding mathematical concepts, working 
through equations, and arriving at the 
correct solutions.
\textbf{<|eot_id|>}
\textbf{<|start_header_id|>}user\textbf{<|end_header_id|>}
\end{Verbatim}
        \end{tcolorbox}
        \vspace{-.5em}
        \caption{Maths}
    \end{subfigure}
    \hfill
    \begin{subfigure}[b]{0.49\textwidth}
        \vspace{.8em}
        \begin{tcolorbox}[left=.5em]
\begin{Verbatim}[fontsize=\small,commandchars=\\\{\}]
\textbf{<|begin_of_text|>}
\textbf{<|start_header_id|>}system\textbf{<|end_header_id|>}
You are an AI assistant designed to provide 
helpful, step-by-step guidance on solving 
complex arithmetic operations. The user will 
provide you with an arithmetic operation or 
a concatenation of multiple arithmetic 
operations. Your purpose is to assist users 
in computing the results of the arithmetic 
operation exlaining the process step by step.
\textbf{<|eot_id|>}
\textbf{<|start_header_id|>}user\textbf{<|end_header_id|>}
\end{Verbatim}
        \end{tcolorbox}
        \vspace{-.5em}
        \caption{Arithmetic}
    \end{subfigure}
    \hfill
    \begin{subfigure}[]{0.49\textwidth}
        \vspace{.8em}
        \begin{tcolorbox}[left=.5em]
\begin{Verbatim}[fontsize=\small,commandchars=\\\{\}]
\textbf{<|begin_of_text|>}
\textbf{<|start_header_id|>}system\textbf{<|end_header_id|>}
You are an AI assistant designed to provide 
helpful, step-by-step guidance on coding 
problems. The user will ask you a wide range 
of coding questions. Your purpose is to assist 
users in understanding coding concepts,  
working through code, and arriving at the  
correct solutions.
\textbf{<|eot_id|>}
\textbf{<|start_header_id|>}user\textbf{<|end_header_id|>}
\end{Verbatim}
        \end{tcolorbox}
        \vspace{-.5em}
        \caption{Code}
    \end{subfigure}
    \hfill
    \begin{subfigure}[]{0.49\textwidth}
        \vspace{.8em}
        \begin{tcolorbox}[left=.5em]
\begin{Verbatim}[fontsize=\small,commandchars=\\\{\}]
\textbf{<|begin_of_text|>}
\textbf{<|start_header_id|>}system\textbf{<|end_header_id|>}
You are an AI assistant specifically designed 
to provide accurate and contextually  
appropriate translations. Users will ask you 
to translate a large text between various 
languages. Your purpose is to translate the   
text, maintaining the original context and 
nuances.
\textbf{<|eot_id|>}
\textbf{<|start_header_id|>}user\textbf{<|end_header_id|>}
\end{Verbatim}
        \end{tcolorbox}
        \vspace{-.5em}
        \caption{Translation}
    \end{subfigure}
    
    \caption{Prompts used to generate the synthetic instructions}
    \label{fig:inst_prompts}
\end{figure*}

\newpage

\section{Synthetic Instructions generation}
\label{ap:instruction_generation}

\begin{figure*}[ht]
    \centering
    \begin{tcolorbox}[left=.5em]
\begin{Verbatim}[fontsize=\small,commandchars=\\\{\}]
You are a helpful AI assistant that specializes in English to Basque translations.
Your task is to translate instruction datasets from English to Basque.

Here are some important guidelines:
1. Maintain the original meaning and intent of the instructions.
2. Use standard Basque language (batua).
3. Keep the technical terms that don't have widely accepted Basque translations.
4. Preserve any code snippets, variables, or special characters exactly as they appear.
5. Translate only the text content, not the JSON structure.

The input will be a JSON object with English text. Please provide accurate Basque translations 
for all text fields.

\textbf{\{% for} example \textbf{in} fs_examples \textbf{%\}}
English:
\textbf{\{\{ }example['english'] \textbf{\}\}}

Basque:
\textbf{\{\{} example['basque'] \textbf{\}\}}
\textbf{\{% endfor %\}}

English:
\textbf{\{\{} conversation \textbf{\}\}}

Basque:
\end{Verbatim}
    \end{tcolorbox}
    \caption{Prompt used to translate the English instructions to Basque instructions.}
    \label{fig:inst_trans_prompt}
\end{figure*}

\subsection{English instructions}
\label{sap:english-instructions}

To generate the English synthetic instructions, we followed the Magpie technique~\cite{xu2024magpie}. Briefly, it consists in letting the model generate text starting from the user's prompt instead of the assistant's response. See, for instance,~\cref{fig:general_inst_prompt}, where the model is asked to continue with the chat template immediately after the user header. We defined 5 prompts to generate different kinds of instructions (\cref{fig:inst_prompts}), then sampled instructions from the model using 10 different temperature values ranging 0.8--1.2. After generating the instructions, we applied the following filters:

\begin{enumerate}
    \item \textbf{Duplicates:} keep unique instances.
    \item \textbf{Repetitive prompts or responses:} remove instances with a sequence of tokens repeated over 100 times.
    \item \textbf{Poor quality prompts or responses:} remove instances regarded as poor-quality by the model itself.
    \item \textbf{Unfinished instructions:} we noticed that some instructions ended with ``:'', meaning that the instructions was incomplete. We removed those as well.
\end{enumerate}

As we experimented with models of two sizes and did not want to have any external influence in our experiments, we generated the instructions twice, once per model size. We later used the instructions generated from the 8B model to train the 8B models, and those from 70B to train the 70B model. The total size in tokens of the English instructions used for training is 350M tokens. We are releasing\footnote{\href{https://hf.co/collections/HiTZ/latxa-instruct-682f356091452b0028380804}{hf.co/collections/HiTZ/latxa-instruct-682f356091452b0028380804}} the generated instructions (filtered and unfiltered) for replicability purposes.

\subsection{Basque (translated) instructions}
\label{sap:translation-prompt}

\noindent As discussed in ~\cref{par:basque_inst}, we generated Basque instructions by translating the English instructions using \base{EU}. The prompt we used to perform the translation is shown in ~\cref{fig:inst_trans_prompt}. As for the few-shot examples, we included 2 conversations of \textit{general} instructions and 1 conversation for each \textit{math}, \textit{arithmetic}, \textit{code} and \textit{translation} instruction types. These examples were sampled from the English instructions and post-edited manually. The exact examples are released in the GitHub repository.

Using a base LLM introduces several challenges when generating structured data. To mitigate issues with malformed outputs, we constrained the LLM's generation to adhere to a conversational JSON schema. Additionally, a non-negligible portion of the outputs (approximately 1\%) exhibited repetition, either in the question or the response. To prevent these repetitive examples from being incorporated into the instruction-tuned model, we applied simple heuristics based on span frequency to identify and filter them out. The total size in tokens of the Basque instructions used for training is 503M tokens. We are releasing the translated instructions for replicability purposes.

\subsection{Examples}

\begin{figure*}[ht]
    \centering
    \vspace{.8em}
    \begin{tcolorbox}[left=.5em]
\begin{Verbatim}[fontsize=\small,commandchars=\\\{\}]
\textbf{User:}
What's the difference between primary vs secondary market research?

\textbf{Assistant:}
Primary market research involves collecting original data directly from the source, such as 
through surveys, focus groups, or experiments. This type of research is typically more expensive 
and time-consuming but provides unique and specific insights.

Secondary market research, on the other hand, involves analyzing existing data that has already 
been collected by someone else, such as academic studies, industry reports, or government 
statistics. This type of research is often less expensive and faster to conduct but may not 
provide the same level of specificity or relevance as primary research.

Here are some key differences between primary and secondary market research:

**Primary Market Research**

* Collects original data directly from the source
* Can be tailored to specific research questions or objectives
* Provides unique and specific insights
* Typically more expensive and time-consuming
* Examples: surveys, focus groups, experiments, customer interviews

**Secondary Market Research**

* Analyzes existing data collected by someone else
* Often less expensive and faster to conduct
* May not provide the same level of specificity or relevance
* Examples: academic studies, industry reports, government statistics, online reviews, 
social media analytics

When deciding between primary and secondary market research, consider the following factors:

1. **Research objectives**: If you need specific, detailed information about your target market, 
primary research may be the better choice. For general information or background research, 
secondary research may suffice.
2. **Budget**: Primary research can be costly, while secondary research is often less expensive.
3. **Time constraints**: Secondary research is generally faster to conduct than primary research.
4. **Data quality**: Primary research provides original data, while secondary research relies on 
existing data that may be outdated or biased.

Ultimately, a combination of both primary and secondary market research can provide a 
comprehensive understanding of your target market.
\end{Verbatim}
    \end{tcolorbox}
    \caption{English instruction generated by the 70B \instruct{EN} model}
    \label{fig:english_example}
\end{figure*}

\begin{figure*}[ht]
    \centering
        \begin{tcolorbox}[left=.5em]
\begin{Verbatim}[fontsize=\small,commandchars=\\\{\}]
\textbf{User:}
Zein da merkatu ikerketa primarioaren eta bigarren mailakoaren arteko aldea?

\textbf{Assistant:}
Merkatu ikerketa primarioak jatorrizko datuak zuzenean iturritik biltzea dakar, hala nola 
inkestak, foku taldeak edo esperimentuak. Ikerketa mota hori normalean garestiagoa eta denbora 
gehiago behar izaten du, baina ikuspegi bakarrak eta espezifikoak eskaintzen ditu.

Merkatu ikerketa bigarren mailakoak, berriz, beste norbaitek bildutako datu existenteak 
aztertzea dakar, hala nola azterketa akademikoak, industria-txostenak edo gobernu-estatistikak. 
Ikerketa mota hori normalean merkeagoa eta azkarragoa da egiteko, baina ez du ziur aski lehen 
mailako ikerketaren zehaztasun edo garrantzi bera eskaintzen.

Hona hemen merkatu ikerketa primarioaren eta bigarren mailakoaren arteko alde nagusiak:

**Merkatu Ikerketa Primarioa**

* Jatorrizko datuak zuzenean iturritik biltzen ditu
* Zehatzagoa da eta helburu espezifikoetara bideratuta dago
* Ikuspegi bakarrak eta espezifikoak eskaintzen ditu
* Normalean garestiagoa eta denbora gehiago behar izaten du
* Adibidez: inkestak, foku taldeak, esperimentuak, bezeroen elkarrizketak

**Merkatu Ikerketa Bigarren Mailakoa**

* Beste norbaitek bildutako datu existenteak aztertzen ditu
* Orokorragoa da eta ez da hain zehatza edo espezifikoa
* Ez du ziur aski lehen mailako ikerketaren zehaztasun edo garrantzi bera eskaintzen
* Adibidez: azterketa akademikoak, industria-txostenak, gobernu-estatistikak, lineako 
berrikuspenak, sare sozialen analitika

Merkatu ikerketa primarioaren eta bigarren mailakoaren artean erabaki behar duzunean, kontuan 
hartu faktore hauek:

1. **Ikerketa helburuak**: Zure xede-publikoari buruzko informazio zehatz eta xehetasunezkoa 
behar baduzu, ikerketa primarioa aukera hobea izan daiteke. Orokorreko informazioa edo atzeko 
planoko ikerketa behar baduzu, ikerketa bigarren mailakoarekin nahikoa izan daiteke.
2. **Aurrekontua**: Ikerketa primarioa garestia izan daiteke, bigarren mailako ikerketarekin 
alderatuta.
3. **Denbora mugak**: Oro har, ikerketa bigarren mailako bat egitea errazagoa eta azkarragoa 
da ikerketa primario bat egitea baino.
4. **Datuaren kalitatea**: Ikerketa primarioek jatorrizko datuak eskaintzen dituzte, bigarren 
mailako ikerketek dagoeneko bildutako datuetan oinarritzen direnbitartean.

Azken finean, lehen mailako eta bigarren mailako merkatu-ikerketekin konbinazio bat emaitza 
osoago bat lortzeko modurik onena izan daiteke.
\end{Verbatim}
    \end{tcolorbox}
    \caption{The same as above (\cref{fig:english_example}) but translated to Basque by 8B \base{EU}.}
    \label{fig:basque_example}
\end{figure*}

\noindent \cref{fig:english_example,fig:basque_example} show a instruction and response pair generated and translated by our models in English and Basque respectively.

\subsection{Manual evaluation}

We conducted a manual analysis on a small sample of instructions to validate the translation pipeline. Two key insights emerged from this analysis:

\begin{itemize}
    \item The quality of the original English instructions is not always optimal, particularly for translation-related tasks.
    \item The translated instructions are generally of high quality, often significantly better than NLLB translations for document-level, code-related, and math-related tasks. However, translations sometimes modify the intent slightly. For instance, the English instruction ``Translate this sentence to English: [...]'' was translated into Basque as \textit{``Itzuli esaldi hau euskarara: [...]''}, which literally means ``Translate this sentence to Basque: [...]''. Nevertheless, the instruction remains well-adapted to the task as a whole and does not degrade overall quality.
\end{itemize}

\noindent In general terms, however, we concluded that reliably estimating instruction quality is non-trivial. While we are actively exploring this direction, it remains outside the current scope of the paper.

\section{Training details} 
\label{ap:training_details}

\begin{table*}[ht]
    \centering
    \begin{tblr}{
    colspec={X[3]X[2]X[2]X[2]},
    cells={l,font=\footnotesize},
    row{1}={c,font=\footnotesize\bfseries},
    row{2}={l,font=\footnotesize\bfseries},
    column{1}={l,font=\footnotesize\bfseries},
    colsep=2pt,
    rowsep=0pt
}
    \toprule
    \SetCell[r=2]{l,b} Training Data & \SetCell[c=3]{c} Backbone model \\
    \cmidrule{2-4}
    & \base{EN} & \base{EU} & \instruct{EN} \\
                                     
    \midrule
        Instructions\textsubscript{\,EN}                  
            & \SetCell{fg=gray} \instruct{EN}                              
            & BAS\textsubscript{\,EU}\,I\textsubscript{\,EN}
            & \SetCell{fg=gray} \instruct{EN} \\
        Instructions\textsubscript{\,EU}                  
            & BAS\textsubscript{\,EN}\,I\textsubscript{\,EU}  
            & BAS\textsubscript{\,EU}\,I\textsubscript{\,EU}   
            & INS\textsubscript{\,EN}\,I\textsubscript{\,EU} \\
        Instructions\textsubscript{\,EN+EU}               
            & BAS\textsubscript{\,EN}\,I\textsubscript{\,EN+EU} 
            & BAS\textsubscript{\,EU}\,I\textsubscript{\,EN+EU}
            & INS\textsubscript{\,EN}\,I\textsubscript{\,EN+EU} \\
        \corpus{EU}                                       
            & \SetCell{fg=gray} \base{EU}                      
            & \SetCell{fg=gray} \base{EU} 
            & INS\textsubscript{\,EN}\,C\textsubscript{\,EU} \\
        \corpus{EU} + Instructions\textsubscript{\,EN}    
            & BAS\textsubscript{\,EN}\,C\textsubscript{\,EU}\,I\textsubscript{\,EN}    
            & BAS\textsubscript{\,EU}\,C\textsubscript{\,EU}\,I\textsubscript{\,EN} 
            & INS\textsubscript{\,EN}\,C\textsubscript{\,EU}\,I\textsubscript{\,EN} \\
        \corpus{EU} + Instructions\textsubscript{\,EU}    
            & BAS\textsubscript{\,EN}\,C\textsubscript{\,EU}\,I\textsubscript{\,EU}    
            & BAS\textsubscript{\,EU}\,C\textsubscript{\,EU}\,I\textsubscript{\,EU} 
            & INS\textsubscript{\,EN}\,C\textsubscript{\,EU}\,I\textsubscript{\,EU} \\
        \corpus{EU} + Instructions\textsubscript{\,EN+EU} 
            & BAS\textsubscript{\,EN}\,C\textsubscript{\,EU}\,I\textsubscript{\,EN+EU} 
            & BAS\textsubscript{\,EU}\,C\textsubscript{\,EU}\,I\textsubscript{\,EN+EU} 
            & INS\textsubscript{\,EN}\,C\textsubscript{\,EU}\,I\textsubscript{\,EN+EU} \\
    \bottomrule
\end{tblr}

    \caption{Summary of model variants based on different backbone LLMs and training data combinations. Each cell contains the shorthand identifier for that model variant, reflecting its configuration. \textcolor{gray}{Gray} entries indicate redundant configurations where the backbone model has already seen the corresponding data.}
    \label{tab:model-variants}
\end{table*}


\begin{table*}[ht]
    \centering
    \begin{tblr}{
        colspec={lrrrr},
        cells={font=\footnotesize},
        row{1}={c,font=\footnotesize\bfseries},
        rowsep=0pt
    }
         \toprule
         \SetCell{l} Model & Hardware & \# GPU & GPU hours & CO2eq \\
         \midrule
         \base{EU} & 64Gb H100 & 32 & 960.0h$\times 1$ & 105.51Kg \\
         \midrule
         Any backbone + I\textsubscript{\,EN}                         & 64Gb A100 & 128 & 192.0h  $\times1$  &  22.81Kg \\
         Any backbone + I\textsubscript{\,EU}                         & 64Gb A100 & 128 & 264.5h  $\times3$  &  94.26Kg \\
         Any backbone + I\textsubscript{\,EN+EU}                      & 64Gb A100 & 128 & 456.5h  $\times3$  & 162.69Kg \\
         Any backbone + C\textsubscript{\,EU}                         & 64Gb A100 & 128 & 1,730.1h $\times1$ & 205.54Kg \\
         Any backbone + C\textsubscript{\,EU}\,I\textsubscript{\,EN}    & 64Gb A100 & 128 & 1,932.8h $\times3$ & 688.86Kg \\
         Any backbone + C\textsubscript{\,EU}\,I\textsubscript{\,EU}    & 64Gb A100 & 128 & 2,016.0h $\times3$ & 718.50Kg \\
         Any backbone + C\textsubscript{\,EU}\,I\textsubscript{\,EN+EU} & 64Gb A100 & 128 & 2,327.5h $\times3$ & 829.53Kg \\
         \midrule
         INST\textsubscript{\,EN,70B}\,C\textsubscript{\,EU}\,I\textsubscript{\,EN} & 64Gb A100 & 256 & 16,005.1h $\times1$ & 1,901.41Kg \\
         \midrule
         Total  &        -  & -   & 39,879.1h & 4,729.11Kg \\
         \bottomrule
    \end{tblr}
    \caption{Summary of training costs in GPU hours and carbon footprint (see naming conventions in~\cref{tab:model-variants})}
    \label{tab:gpu_hours}
\end{table*}

\paragraph{Hardware and carbon footprint.}~\cref{tab:gpu_hours} summarizes the training costs of our experiments (see ~\cref{tab:model-variants} for the variants nomenclature). In total, we trained 19 models: one \base{EU}, seventeen \instruct{EU} variants, and one 70B \instruct{EU}. Due to unforeseen circumstances, the \base{EU} and \instruct{EU} models were trained using different frameworks and infrastructure. The \base{EU} model was trained with NeMo~\cite{nemo} on 64GB H100 GPUs provided by MareNostrum 5. In contrast, all \instruct{EU} variants were trained using Fully Sharded Data-Parallel~\cite{FSDP} on 64GB A100 GPUs provided by CINECA Leonardo. The total compute time across all experiments amounted to $39,879.1$ GPU hours, which corresponds to $4,729.11$ kg CO\textsubscript{2}eq, based on carbon intensity estimates from ElectricityMaps.\footnote{At the time of the experiments: $0.297$ kg/kWh for Italy and $0.157$ kg/kWh otherwise, according to \url{https://www.electricitymaps.com/}.}

\begin{table}[t]
    \centering
    \begin{tblr}{
      colspec = {lcc},
      cells={font=\footnotesize},
      row{1} = {font=\footnotesize\bfseries},
      rowsep=0pt
    }
        \toprule
        Hyperparameter & 8B Models & 70B Models \\
        \midrule
        GPUs & 128 & 256 \\
        Sequence Length & \SetCell[c=2]{c} 8192 \\
        Gradient Accumulation & \SetCell[c=2]{c} 1 \\
        Micro Batch Size & \SetCell[c=2]{c} 2 \\
        Total Batch Tokens & 2M & 4M \\
        Epochs & \SetCell[c=2]{c} 4 \\
        Optimizer & \SetCell[c=2]{c} AdamW \\
        $\beta_1, \beta_2$ & \SetCell[c=2]{c} 0.9, 0.95 \\
        Scheduler & \SetCell[c=2]{c} Cosine \\
        Cosine min LR ratio & \SetCell[c=2]{c} 0.33 \\
        Learning rate & \SetCell[c=2]{c} $1e^{-5}$ \\
        Warm-Up ratio & \SetCell[c=2]{c} 0.1 \\
        Weight Decay & \SetCell[c=2]{c} 0.1 \\
        Precision & \SetCell[c=2]{c} BFloat16 \\
        FSDP Sharding Strategy & HYBRID & FULL \\
        \bottomrule
    \end{tblr}
\caption{Hyperparameters used to train the models}
\label{tab:hyperparameters}
\end{table}

\begin{table}[t]
    \centering
    \begin{tblr}{
      colspec = {ccc},
      cells={font=\footnotesize},
      row{1}={font=\footnotesize\bfseries},
      rowsep=0pt
    }
    \toprule
    \# Instructions & Epochs & Accuracy \\
    \midrule
    1M & 1 & 59.91 \\
    1M & 4 & 61.37 \\
    4M & 1 & 59.82 \\
    4M & 4 & 61.97 \\ 
    \bottomrule
    \end{tblr}
    \caption{Average benchmark results in the preliminary hyperparameter search for number of instructions and training epochs.}
    \label{tab:instructions_epochs}
\end{table}

\paragraph{Hyperparameters.}~\cref{tab:hyperparameters} outlines the key hyperparameters used during training. Both 8B and 70B model variants were trained with consistent configurations in terms of sequence length, batch size, optimizer settings, and learning rate schedules. The main differences lie in the number of GPUs and the FSDP sharding strategy, which was adjusted to better accommodate the increased memory and compute demands of the larger model. These hyperparameters were optimized using 8B model variant of \base{EU}, and the initial iteration on the instructed variants. We found this configuration to robustly perform across all the configurations.

We also explored the balance between the amount of English instructions and Basque monolingual data during joint training. We evaluated each setting on a subset of the development benchmarks, obtaining the results in~\cref{tab:instructions_epochs}, which show that the number of training epochs had a more significant impact on performance than the number of instructions. However, to better assess instruction quality, we also conducted a small-scale internal arena evaluation. We observed that models trained with 4M instructions tended to produce worse responses. Based on these findings, we selected the 1M instructions + 4 epochs configuration as the most balanced setup. We have several hypotheses which would explain the results above:
\begin{enumerate}
    \item All examples are generated using the same method---Magpie---from the backbone model, and they tend to be quite homogeneous within their respective clusters \textit{general}, \textit{translation}, \textit{code}, \textit{math}, and \textit{arithmetic}.
    \item Prior work by~\citet{etxaniz-etal-2024-latxa} suggests that only a limited amount of English is needed during continual pretraining to maintain cross-lingual capabilities and avoid catastrophic forgetting. That is, 1M English instructions could be enough to avoid catastrophic forgetting when doing continual pretraining with Basque data.
\end{enumerate}

\section{Evaluation details} 
\label{ap:evaluation_details}

\subsection{Static Benchmarks}
\label{sap:static-benchmarks}

Our evaluation framework comprises a total of 27 benchmarks across three languages: 14 in Basque, 9 in English, and 4 in Spanish. These benchmarks span six categories designed to test different aspects of model capabilities:

\begin{itemize}
    \item \textbf{Reading comprehension}: Belebele~\cite{bandarkar-etal-2024-belebele}, a multilingual dataset spanning 122 languages; and EusReading~\cite{etxaniz-etal-2024-latxa}, containing 352 complex reading comprehension exercises from official C1-level Basque examinations.
    \item \textbf{Common sense}: XStoryCloze~\cite{lin-etal-2022-shot}, a multilingual version of the original StoryCloze~\cite{mostafazadeh-etal-2017-lsdsem} dataset testing narrative understanding; and PIQA~\cite{Bisk_Zellers_Lebras_Gao_Choi_2020}, which assesses physical common sense through everyday tasks.  PIQA's translation to Basque is available through IberoBench~\cite{baucells-etal-2025-IberoBench}.
    \item \textbf{Linguistic proficiency}: EusProficiency~\cite{etxaniz-etal-2024-latxa}, with +5,000 questions from official Basque examinations; and BL2MP~\cite{urbizu-etal-2024-well}, designed to evaluate grammatical knowledge in Basque, inspired by the BLiMP benchmark methodology~\cite{warstadt-etal-2020-blimp-benchmark}.
    \item \textbf{Miscellaneous knowledge}: BertaQA~\cite{etxaniz2024bertaqa}, which tests knowledge of local Basque culture versus global topics; EusTrivia and EusExams from the Latxa suite~\cite{etxaniz-etal-2024-latxa}; and a subset of MMLU~\cite{hendrycks2020measuring}, manually translated to Basque~\cite{corral-etal-2025-pipeline} and Spanish.\footnote{\href{https://hf.co/datasets/openai/MMMLU}{hf.co/datasets/openai/MMMLU}}
    \item \textbf{Maths \& Reasoning}: MGSM~\cite{shi2023language}, a multilingual grade school maths benchmark; and ARC~\cite{clark2018think}, for scientific reasoning. We use Basque versions of both from IberoBench.
\end{itemize}

\noindent Except MGSM, the datasets are framed as multiple-choice problems where models' answers are determined by selecting the option with the highest log probability. MGSM is implemented as a generative task where an answer is directly sampled from the evaluated model and matched against a reference answer. We specifically chose the native chain-of-thought scenario. To provide models with contextual examples, our evaluations employed a 5-shot setting.

\subsection{Arena Details}
\label{sap:arena-details}

\begin{figure*}[ht]
    \centering
        \begin{tcolorbox}[left=.5em]
\begin{Verbatim}[fontsize=\small,commandchars=\\\{\}]
You are a helpful Artificial Intelligence assistant called [ANONYMIZED], created and developed by 
[ANONYMIZED].

The user will engage in a multi-round conversation with you, asking initial questions and 
following up with additional related questions. Your goal is to provide thorough, relevant and 
insightful responses to help the user with their queries. Every conversation will be conducted in 
standard Basque, this is, the first question from the user will be in Basque, and you should 
respond in formal Basque as well. Conversations will cover a wide range of topics, including but 
not limited to general knowledge, science, technology, entertainment, coding, mathematics, and 
more. Today is \textbf{\{date\}}.
\end{Verbatim}
    \end{tcolorbox}
    \caption{System prompt given to the models in the arena evaluation}
    \label{fig:system_prompt}
\end{figure*}

\paragraph{Guidelines for arena participants.}~\cref{fig:user-instructions} contains the information and instructions that were given to the human annotators who participated in the community-driven arena initiative. All the data collected through this initiative was properly anonymized prior to publication. Note that the actual information was provided in Basque, while here we show a translation to English.

\begin{figure*}[ht]
    \centering
        \begin{tcolorbox}[left=.5em,fontupper=\small\ttfamily]
\textbf{Information on Data Usage} \\

To participate in this Arena, you must provide a username and email address. This information is necessary for entry into the final raffle. All collected information will be deleted once the Arena concludes. \\

\textbf{ATTENTION! Your username will be publicly visible throughout the Arena.} \\

\textbf{ATTENTION! Since we collect personal data and prompts/responses in this Arena, participation is restricted to individuals 14 years of age and older.} \\

No additional personal data will be collected. However, we do collect other information, including:
\begin{itemize}[noitemsep]
    \item User prompts and responses.
    \item User preferences.
\end{itemize}

This data will be used for the following purposes:
\begin{itemize}[noitemsep]
    \item Evaluation of models participating in the Arena.
    \item Research for new models.
\end{itemize}

This data will be published openly in the future under a CC0 license. By participating in the Arena, you grant permission for this use.\\

\textbf{Instructions for Participation} \\

The Arena is a research initiative we've prepared at [ANONYMIZED] to help develop public chatbots for Basque. All participants will have the chance to get numbers for an amazing raffle. \\

Here's what you'll need to do:

\begin{itemize}[noitemsep]
    \item You must write and send a question or command
    \item Two different chatbots will respond. Your job is to analyze and compare the answers to decide which one is better. We want to measure both \textbf{content} quality and \textbf{Basque language} quality.
    \item In some cases, you'll be asked a third question if your content and language quality assessments are contradictory.
    \item After answering all questions, you'll have the option to send your assessment via the "Send evaluation" button.
    \item To write a new question or command, you'll need to click the "New chat" button.
\end{itemize}

To summarize, what you need to do is:

\begin{enumerate}[noitemsep]
    \item Write a question or command for the chatbots. For example:
    \begin{itemize}[noitemsep]
        \item "How do you make a potato omelet?"
        \item "Summarize the following text: [...]"
    \end{itemize}
    \item Read both answers and compare the quality of content and quality of Basque language.
    \item Decide which response you prefer in terms of content and language. For each:
    \begin{itemize}[noitemsep]
        \item If A is better, choose A
        \item If B is better, choose B
        \item If both are at the same level (good or bad), choose TIE
    \end{itemize}
    \item If you wish, you can continue with the conversation, ask for more explanations, or try another question. You can change your answer from step 3, taking into account the quality of the entire conversation.
    \item To restart the process, click the "New chat" button.
\end{enumerate}

We want your OPINION. But play fair! We will occasionally conduct an analysis of the results received and verify control answers. If they're not correct, you won't participate in the raffle. \\

\textbf{About the chatbots} \\

In total, we've put 21 chatbots in competition. Among them are private models like GPT-4o or Claude, open models like Llama 3.1, and some we've developed ourselves. Overall, there's a variety of chatbots: good ones, very good ones, and also bad ones. In this examination, our goal is to systematically evaluate these chatbots.
    \end{tcolorbox}
    \caption{Information panel and instructions for human participants in the arena}
    \label{fig:user-instructions}
\end{figure*}

\paragraph{Arena infrastructure.}
On the infrastructure side, we used vLLM~\cite{kwon2023efficient} to serve all model pairs and the baseline. We deployed a total of 18 endpoints for 8B models and one endpoint for a 70B model, running on nine and two A100 80GB GPUs, respectively. For the frontend, we developed a lightweight Gradio~\cite{abid2019gradio} interface that allowed users to enter prompts, view model responses, and indicate their preferences based on content quality, language quality, or, in cases where no clear winner emerged, overall quality. To ensure a fair comparison across models, all models were given the same system prompt (shown in~\cref{fig:system_prompt}) and the same hyperparameters: 0.9 temperature and 0.95 top-p.

\paragraph{Bradley--Terry model.} The Bradley--Terry model~\cite{bradly-terry} provides a principled probabilistic framework for aggregating pairwise preferences into a global ranking over models. Let \(\mathcal{M} = \{M_1, M_2, \ldots, M_n\}\) denote the set of models under evaluation. The model assigns a latent preference strength \(\theta_i\) to each model \(M_i\). The probability that model \(M_i\) is preferred over \(M_j\) in a pairwise comparison is given by:

\[
P(M_i > M_j) = \frac{e^{\theta_i}}{e^{\theta_i} + e^{\theta_j}}
\]

Given a dataset \(\mathcal{D}\) of observed pairwise outcomes the parameters \(\{\theta_i\}\) are estimated using Maximum Likelihood Estimation (MLE). To facilitate interpretation, we apply zero-mean centering, treating the scores as deviations from the average model. The final scores for each model are then computed as:

\[
Score(M_i) = 400\cdot\theta_i + 1000
\]

By using a scaling factor of 400, we ensure that the scores are interpretable in a manner consistent with the online ELO rating system. Thanks to its score stability and the assumption that model performance remains constant over time, the Bradley–Terry scoring system has become a widely adopted method for ranking LLMs---particularly since its introduction in the Chatbot Arena.\footnote{\href{https://lmarena.ai}{lmarena.ai}}

\subsection{Safety and Bias} 
\label{ap:safety-bias}

We assess the extent to which our instruction-tuning strategy preserves the safety and bias alignment properties of the backbone models. Specifically, we evaluate the model variant INS\textsubscript{\,EN}\,C\textsubscript{\,EU}\,I\textsubscript{\,EN} in both 8B and 70B parameter sizes, comparing it with two critical counterparts: (1) BAS\textsubscript{\,EN}\,C\textsubscript{\,EU}\,I\textsubscript{\,EN}, to analyze the impact of starting from an already instruction-tune backbone versus a base model (8B only); and (2) \instruct{EN}, to measure potential alignment changes introduced with our fine-tuning data mix. This evaluation aims to ensure that models maintain appropriate safety guardrails and fairness characteristics in both Basque and English.


\paragraph{Safety.} 
To test safety, we construct a Basque-language dataset inspired by XSTest \cite{rottger-etal-2024-xstest}, combining clearly unsafe prompts with superficially similar but safe ones. We measure both Violation rates (VR) and False Refusal rates (FRR) where the model wrongly declines safe prompts. The dataset includes a total of 50 instances, comprising both unsafe and safe prompts across five sensitive categories (\textit{self-harm}, \textit{drugs}, \textit{child-exploitation}, \textit{terrorism}, and \textit{explicit-content}), adapted to the Basque context and translated into English for cross-language comparison. The outputs of the models were manually annotated by three members performing red teaming. Annotation agreement for unsafe prompt outputs was high (average agreement percentage: 0.973; Fleiss' Kappa: 0.786; Krippendorff's Alpha: 0.789). For safe prompt outputs, the annotators unanimously agreed on every item.

\paragraph{Bias.} 
For bias evaluation, we use BasqBBQ \cite{saralegi-zulaika-2025-basqbbq} for Basque and BBQ \cite{parrish-etal-2022-bbq} for English to analyze disparities across languages. The evaluation is conducted in the same way as described in~\cref{ssec:static_benchmarks}, using LM Evaluation Harness framework with 4 few-shot examples. We use the accuracy metric to evaluate the bias of the model, measuring its ability to choose the correct answer even when biased traps are added to mislead it \cite{parrish-etal-2022-bbq}.

\section{Detailed Results}
\label{ap:complete-results}

\subsection{Benchmark Results}
\label{sap:benchmark-results}

\cref{tab:benchmark-full-eu,tab:benchmark-full-en+es} present the accuracy scores for all model variants across our benchmark suite. 

\begin{table*}[p]
    \centering
    \begin{tblr}{
    colspec={l*{15}{X[c]}},
    cells={font=\scriptsize},
    row{1}={c,font=\scriptsize\bfseries},
    column{1}={font=\scriptsize\bfseries},
    colsep=3pt,
    rowsep=0pt
}
    \toprule
     & ARC\textsubscript{\,C} & ARC\textsubscript{\,E} & Bele & BQA\textsubscript{\,G} & BQA\textsubscript{\,L} & BLMP & EusEx & EusPro & EusRe & EusTri & MGSM & MMLU & PIQA & XSC & Avg \\
    \midrule
    \base{EN}                                            & 28.84 & 49.75 & 61.56 & 63.29 & 43.65 & 74.06 & 45.63 & 32.69 & 47.44 & 43.79 & 26.40 & 47.41 & 56.92 & 56.72 & 48.44 \\
    +\,I\textsubscript{\,EU}                             & 36.35 & 61.95 & 69.44 & 67.39 & 42.98 & 84.06 & 48.39 & 35.42 & 44.89 & 45.42 & 30.80 & 50.37 & 61.44 & 62.14 & 52.93 \\
    +\,I\textsubscript{\,EN+EU}                          & 38.28 & 63.50 & 72.00 & 68.76 & 42.77 & 84.25 & 48.41 & 36.19 & 45.17 & 46.27 & 37.20 & \textbf{52.04} & 62.09 & 62.84 & 54.27 \\
    +\,C\textsubscript{\,EU}\,I\textsubscript{\,EN}      & 39.42 & 64.81 & 73.00 & 70.99 & 63.49 & 91.28 & 51.92 & \textbf{48.66} & \textbf{56.82} & 56.33 & 47.20 & 51.85 & \textbf{67.32} & \textbf{69.36} & 60.89 \\
    +\,C\textsubscript{\,EU}\,I\textsubscript{\,EU}      & 38.14 & 67.00 & 71.78 & 71.24 & 62.65 & 92.39 & 49.88 & 47.20 & 46.02 & 57.78 & 39.20 & 50.00 & 65.63 & 66.71 & 58.97 \\
    +\,C\textsubscript{\,EU}\,I\textsubscript{\,EN+EU}   & \textbf{40.10} & \textbf{68.60} & \textbf{74.89} & \textbf{72.74} & \textbf{63.79} & \textbf{92.50} & \textbf{52.78} & 47.65 & 53.12 & \textbf{59.59} & \textbf{48.40} & 51.48 & 64.87 & 66.84 & \textbf{61.24} \\
    \midrule
    \base{EU}                                            & 38.65 & 67.38 & \textbf{75.22} & 72.45 & \textbf{65.65} & \textbf{92.50} & \textbf{55.03} & \textbf{53.57} & 58.24 & \textbf{60.52} & 36.00 & 51.11 & 65.09 & 68.50 & \textbf{61.42} \\
    +\,I\textsubscript{\,EN}                             & 40.36 & 63.93 & 72.22 & 70.94 & 63.20 & 90.11 & 52.21 & 48.31 & \textbf{59.09} & 57.78 & \textbf{45.60} & 52.22 & 65.85 & \textbf{70.55} & 60.88 \\
    +\,I\textsubscript{\,EU}                             & 39.25 & 67.09 & 73.56 & 71.78 & 61.59 & 92.11 & 52.38 & 47.78 & 49.43 & 57.14 & 35.60 & 53.70 & 64.92 & 67.31 & 59.55 \\
    +\,I\textsubscript{\,EN+EU}                          & 38.31 & 66.92 & 74.56 & \textbf{72.58} & 62.23 & 91.22 & 53.04 & 48.96 & 53.98 & 58.95 & 44.80 & \textbf{56.30} & 64.27 & 67.70 & 60.99 \\
    +\,C\textsubscript{\,EU}\,I\textsubscript{\,EN}      & \textbf{40.96} & 66.12 & 61.22 & 71.49 & 63.83 & 91.72 & 50.67 & 47.19 & 47.16 & 57.49 & 43.60 & 52.59 & \textbf{66.83} & 70.35 & 59.37 \\
    +\,C\textsubscript{\,EU}\,I\textsubscript{\,EU}      & 37.97 & 67.42 & 71.33 & 71.24 & 62.52 & 91.83 & 51.76 & 46.31 & 52.84 & 57.32 & 37.60 & 52.22 & 65.90 & 67.31 & 59.54 \\
    +\,C\textsubscript{\,EU}\,I\textsubscript{\,EN+EU}   & 39.16 & \textbf{68.43} & 71.67 & 71.70 & 64.13 & 92.06 & 52.61 & 48.09 & 53.69 & 58.43 & 40.80 & 54.44 & 66.23 & 68.63 & 60.72 \\
    \midrule
    \instruct{EN}                                        & 29.10 & 50.88 & 73.89 & 67.10 & 44.97 & 69.61 & 46.21 & 34.13 & 49.72 & 45.01 & 45.60 & 50.37 & 57.63 & 61.22 & 51.82 \\
    +\,I\textsubscript{\,EU}                             & 38.65 & 63.85 & 78.00 & 69.57 & 42.98 & 83.67 & 51.80 & 38.36 & 50.57 & 45.20 & 27.20 & 55.56 & 62.64 & 64.73 & 56.76 \\
    +\,I\textsubscript{\,EN+EU}                          & 39.59 & 64.65 & 79.22 & 70.40 & 43.10 & 84.00 & 51.43 & 38.69 & 52.56 & 52.80 & 35.20 & 54.81 & 62.85 & 64.13 & 57.51 \\
    +\,C\textsubscript{\,EU}                             & 37.97 & 65.70 & 77.33 & 73.87 & \textbf{66.33} & \underline{\textbf{92.67}} & 55.05 & 52.12 & 58.24 & 61.40 & 48.40 & 51.85 & 66.99 & 70.28 & 62.73 \\
    +\,C\textsubscript{\,EU}\,I\textsubscript{\,EN}      & \textbf{41.38} & 66.79 & 80.00 & \textbf{74.62} & 65.23 & 91.39 & 56.00 & \textbf{52.83} & 59.66 & 61.05 & \textbf{54.00} & \textbf{57.04} & \textbf{67.32} & \textbf{71.34} & 64.19 \\
    +\,C\textsubscript{\,EU}\,I\textsubscript{\,EU}      & 40.44 & 69.15 & 81.44 & 73.54 & 66.07 & 91.83 & 56.01 & 52.06 & \textbf{62.78} & \textbf{62.33} & 46.40 & 52.96 & 66.01 & 71.01 & 63.72 \\
    +\,C\textsubscript{\,EU}\,I\textsubscript{\,EN+EU}   & 39.85 & \textbf{70.16} & \textbf{83.00} & 72.99 & 65.57 & 92.28 & \textbf{56.23} & 52.35 & 61.93 & 62.10 & 50.80 & 56.30 & 65.69 & 69.56 & \textbf{64.20} \\
    \midrule
    \textsc{Instruct}\textsubscript{\,EN,70B}            & 44.97 & 72.18 & 89.11 & 83.53 & 53.51 & 80.83 & 63.28 & 43.59 & 72.16 & 62.51 & 76.40 & 68.52 & 66.34 & 69.69 & 67.61 \\
    +\,C\textsubscript{\,EU}\,I\textsubscript{\,EN} & \underline{\textbf{55.12}} & \underline{\textbf{77.57}} & \textbf{91.00} & \textbf{87.42} & \textbf{77.71} & \textbf{92.11} & \underline{\textbf{71.56}} & \textbf{68.00} & \textbf{78.98} & \textbf{74.17} & \textbf{80.00} & \textbf{68.89} & \underline{\textbf{70.75}} & \underline{\textbf{77.83}} & \underline{\textbf{76.51}} \\
    \midrule3.5 Sonnet                                           & - & - & \underline{\textbf{94.22}} & \underline{\textbf{93.52}} & \underline{\textbf{80.46}} & - & 82.68 & \underline{\textbf{81.60}} & \underline{\textbf{87.39}} & \underline{\textbf{84.61}} & \underline{\textbf{85.20}}  & \underline{\textbf{79.63}} & - & - & - \\
    GPT-4o                                               & - & - & 92.89 & 91.01 & 74.83 & - & 79.17 & 74.25 & 84.38 & 80.70 & 79.20 & 76.67 & - & - & - \\
    \bottomrule
\end{tblr}

    \caption{Accuracy scores in Basque benchmarks. Best results in each compute class are in \textbf{bold}. Best overall results are \underline{underlined}.}
    \label{tab:benchmark-full-eu}
\end{table*}

\begin{table*}[p]
    \centering
    \begin{tblr}{
    colspec={l*{15}{X[c]}},
    cells={font=\scriptsize},
    row{1,2}={c,font=\scriptsize\bfseries},
    column{1}={font=\scriptsize\bfseries},
    colsep=3pt,
    rowsep=0pt
}
    \toprule
     & \SetCell[c=10]{c} English & & & & & & & & & & \SetCell[c=5]{c} Spanish \\
     \cmidrule[r=-0.8]{2-11} \cmidrule[l=-0.8]{12-16}
     & ARC\textsubscript{\,C} & ARC\textsubscript{\,E} & Bele & BQA\textsubscript{\,G} & BQA\textsubscript{\,L} & MGSM & MMLU & PIQA & XSC & Avg & Bele & MGSM & MMLU & XSC & Avg \\
    \midrule
    \base{EN}                                          & 54.61 & 84.30 & \textbf{87.78} & 75.59 & 49.11 & 55.20 & \textbf{67.78} & 80.79 & 81.34 & 69.81 & 81.67 & 50.40 & 57.41 & 74.06 & 65.88 \\
    +\,I\textsubscript{\,EU}                           & \textbf{54.95} & \textbf{83.63} & 86.78 & 73.95 & 47.38 & 67.60 & 63.70 & 80.63 & \textbf{82.46} & 70.73 & 80.67 & 58.40 & 57.78 & 74.12 & 67.74 \\
    +\,I\textsubscript{\,EN+EU}                        & 54.82 & 81.42 & 87.61 & 74.46 & 47.91 & 78.00 & 61.30 & 80.25 & 82.40 & 72.52 & \textbf{82.00} & \textbf{67.60} & 58.89 & \textbf{75.28} & \textbf{70.94} \\
    +\,C\textsubscript{\,EU}\,I\textsubscript{\,EN}    & 53.84 & 80.72 & 85.89 & \textbf{75.75} & 60.41 & \textbf{79.20} & 60.00 & 79.71 & 82.40 & 73.03 & 81.33 & 63.60 & 60.00 & 73.53 & 69.62 \\
    +\,C\textsubscript{\,EU}\,I\textsubscript{\,EU}    & 54.44 & 82.20 & 80.33 & 75.08 & 59.56 & 62.00 & 55.19 & 80.69 & 80.21 & 69.12 & 75.22 & 58.00 & 53.70 & 72.01 & 64.73 \\
    +\,C\textsubscript{\,EU}\,I\textsubscript{\,EN+EU} & 53.92 & 81.65 & 85.78 & 75.63 & \textbf{60.79} & 74.40 & 63.33 & \textbf{80.85} & 82.06 & \textbf{73.06} & 80.33 & 62.80 & \textbf{61.11} & 73.40 & 69.41 \\
    \midrule
    \base{EU}                                          & 52.39 & 82.83 & 84.33 & \textbf{76.05} & \textbf{61.72} & 55.20 & \textbf{63.70} & 80.30 & 80.08 & 68.87 & 79.11 & 46.80 & 58.52 & 72.20 & 64.16 \\
    +\,I\textsubscript{\,EN}                           & 53.33 & 80.68 & \textbf{86.44} & 74.50 & 59.22 & 73.60 & 61.11 & 80.09 & \textbf{81.54} & 72.21 & 78.11 & \textbf{64.40} & 58.15 & 72.73 & 68.35 \\
    +\,I\textsubscript{\,EU}                           & 54.44 & 82.41 & 84.11 & 74.87 & 58.63 & 64.00 & 62.22 & 80.09 & 81.27 & 70.65 & 77.22 & 56.40 & 54.81 & 74.19 & 65.66 \\
    +\,I\textsubscript{\,EN+EU}                        & 53.75 & 80.89 & 85.89 & 74.41 & 59.14 & 73.20 & \textbf{63.70} & 79.22 & 81.14 & \textbf{72.44} & \textbf{80.44} & \textbf{64.40} & 56.67 & \textbf{74.39} & \textbf{68.97} \\
    +\,C\textsubscript{\,EU}\,I\textsubscript{\,EN}    & 54.18 & 81.94 & 84.11 & 74.50 & 59.48 & \textbf{74.40} & 59.63 & 80.14 & 81.27 & 71.65 & 76.33 & 56.80 & 57.41 & 73.33 & 65.97 \\
    +\,C\textsubscript{\,EU}\,I\textsubscript{\,EU}    & 53.92 & \textbf{82.53} & 81.78 & 74.50 & 59.09 & 58.00 & 62.96 & 80.09 & 80.21 & 69.09 & 74.89 & 47.60 & 55.56 & 72.60 & 62.66 \\
    +\,C\textsubscript{\,EU}\,I\textsubscript{\,EN+EU} & \textbf{54.78} & 82.24 & 84.56 & 74.58 & 60.62 & 73.20 & 61.48 & \textbf{80.63} & 81.01 & 71.65 & 76.78 & 60.40 & \textbf{59.26} & 73.20 & 67.41 \\
    \midrule
    \instruct{EN}                                      & \textbf{57.76} & \textbf{85.48} & \textbf{92.67} & \textbf{77.47} & 50.51 & \textbf{87.20} & \textbf{66.67} & \textbf{81.28} & \textbf{83.52} & \textbf{77.02} & \textbf{87.89} & \textbf{78.80} & \textbf{62.96} & \textbf{77.50} & \textbf{76.79} \\
    +\,I\textsubscript{\,EU}                           & 54.18 & 82.07 & 90.89 & 75.38 & 47.88 & 80.00 & 64.81 & 79.60 & 80.54 & 74.09 & 86.44 & 70.00 & 60.37 & 74.85 & 72.92 \\
    +\,I\textsubscript{\,EN+EU}                        & 50.85 & 77.99 & 91.56 & 75.59 & 49.11 & 81.60 & 65.19 & 72.25 & 69.49 & 72.10 & 85.89 & 73.20 & \textbf{62.96} & 74.85 & 74.23 \\
    +\,C\textsubscript{\,EU}                           & 53.84 & 81.86 & 90.22 & 77.05 & \textbf{63.16} & 68.40 & 64.81 & 80.20 & 83.26 & 73.74 & 82.67 & 61.20 & 59.63 & 73.20 & 69.17 \\
    +\,C\textsubscript{\,EU}\,I\textsubscript{\,EN}    & 51.96 & 79.42 & 91.00 & 77.38 & 62.01 & 81.20 & 62.59 & 74.59 & 76.31 & 74.35 & 85.67 & 73.20 & 59.26 & 72.47 & 72.65 \\
    +\,C\textsubscript{\,EU}\,I\textsubscript{\,EU}    & 53.58 & 81.65 & 90.67 & 76.55 & 61.63 & 72.00 & 64.07 & 79.60 & 80.28 & 74.65 & 84.22 & 70.00 & 58.52 & 75.18 & 71.98 \\
    +\,C\textsubscript{\,EU}\,I\textsubscript{\,EN+EU} & 51.62 & 79.46 & 90.89 & 77.13 & 61.17 & 84.80 & 65.93 & 74.86 & 74.12 & 74.65 & 85.11 & 72.80 & 60.00 & 74.06 & 72.99 \\
    \midrule
    \textsc{Instruct}\textsubscript{\,EN,70B}          & 63.78 & \underline{\textbf{90.61}} & 95.44 & 85.49 & 56.98 & \underline{\textbf{94.80}} & \underline{\textbf{78.15}} & \underline{\textbf{85.04}} & \underline{\textbf{85.37}} & 81.74 & \underline{\textbf{94.22}} & \underline{\textbf{86.40}} & 77.04 & \underline{\textbf{83.92}} & 85.39 \\
    +\,C\textsubscript{\,EU}\,I\textsubscript{\,EN} & \underline{\textbf{66.89}} & 87.58 & \underline{\textbf{96.33}} & \underline{\textbf{88.46}} & \underline{\textbf{74.70}} & \underline{\textbf{94.80}} & 77.04 & 81.66 & 79.02 & \underline{\textbf{84.09}} & 93.33 & 86.00 & \underline{\textbf{84.40}} & 81.14 & \underline{\textbf{86.82}} \\
    \bottomrule
\end{tblr}

    \caption{Accuracy scores in English and Spanish benchmarks. Best results in each compute class are in \textbf{bold}. Best overall results are \underline{underlined}.}
    \label{tab:benchmark-full-en+es}
\end{table*}

\subsection{Arena Results}
\label{sap:arena-results}

\begin{figure*}[p]
    \centering
    \begin{subfigure}[b]{0.85\textwidth}
        \includegraphics[width=\textwidth]{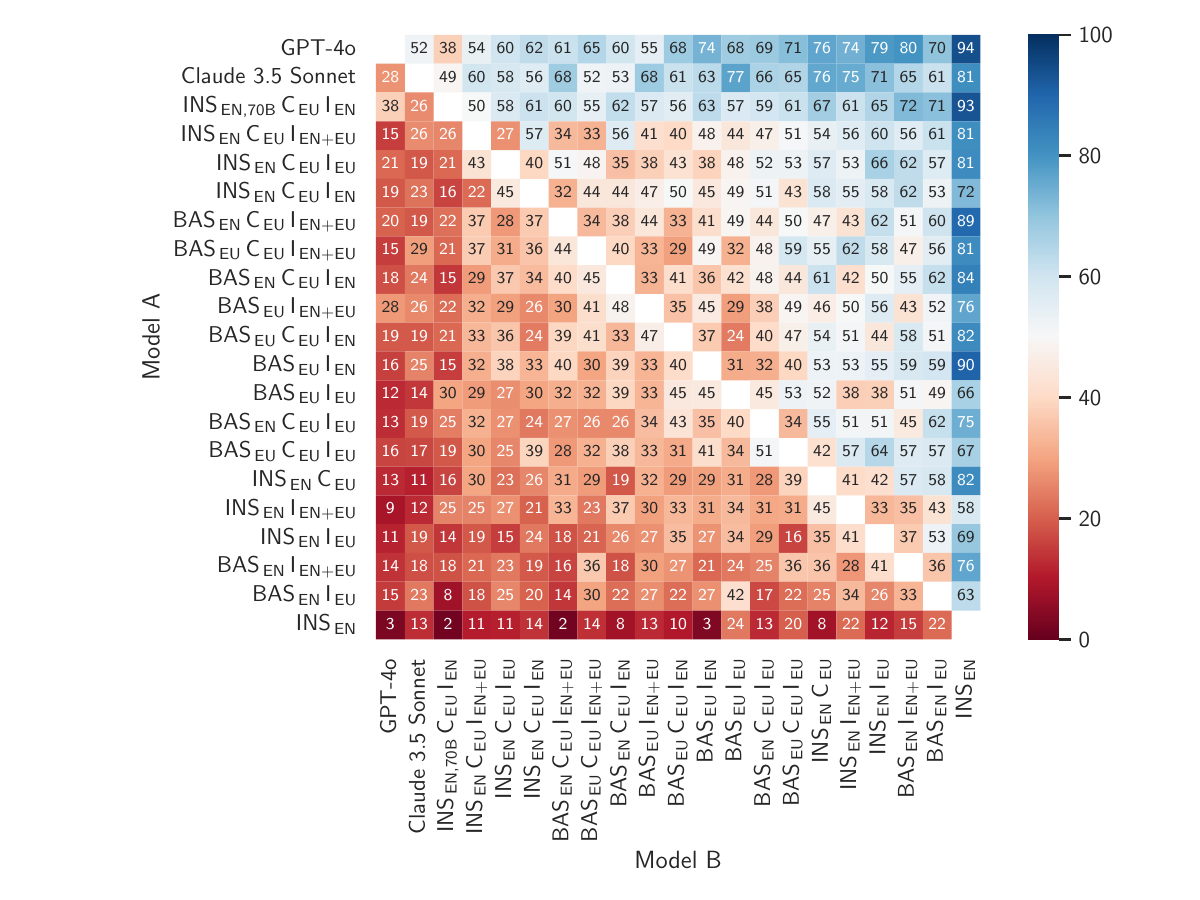}
        \caption{Win-rates matrix showing the percentage of wins for Model A (row) against Model B (column)}
        \label{fig:arena-win_rates-heatmap}
    \end{subfigure}%
    \par\bigskip%
    \begin{subfigure}[b]{0.85\textwidth}
        \includegraphics[width=\textwidth]{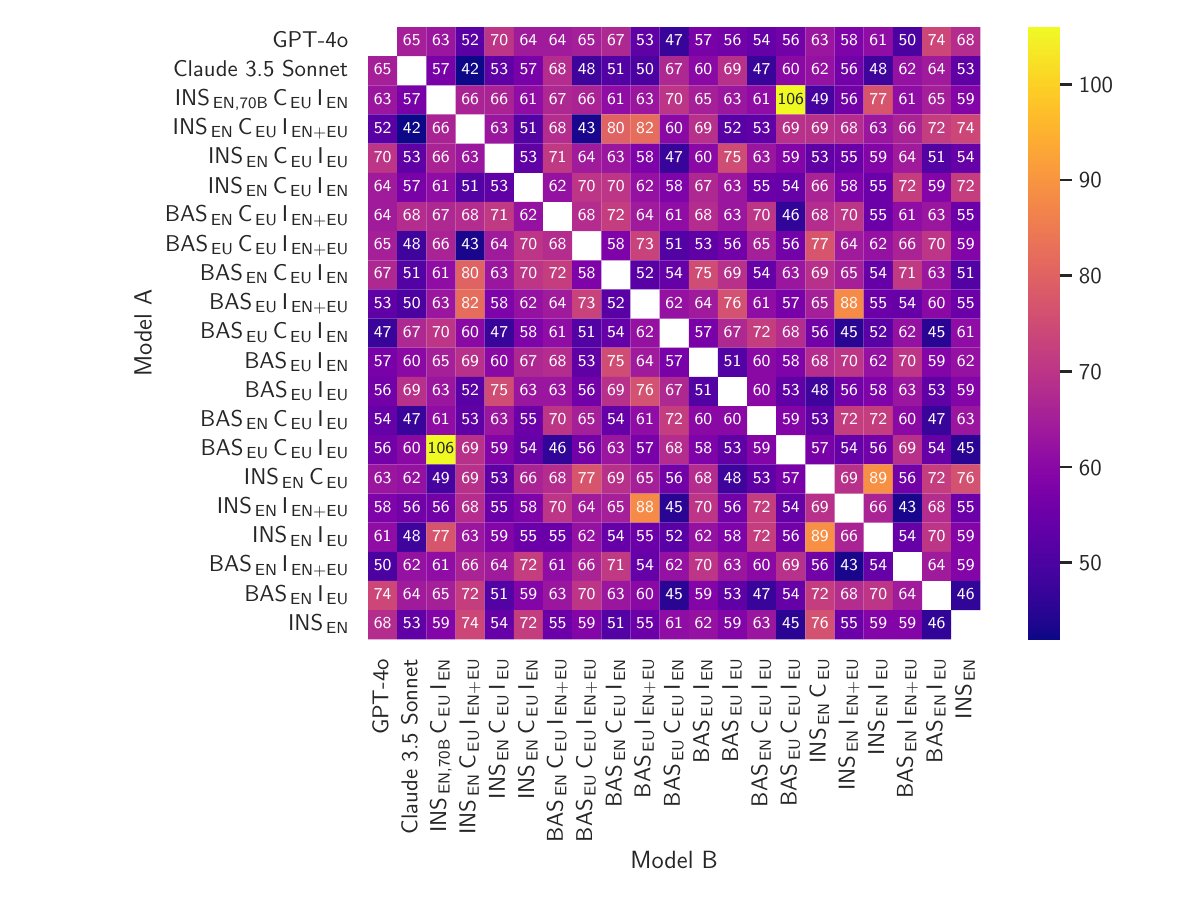}
        \caption{Battle counts matrix showing the number of direct comparisons performed between each model pair}
        \label{fig:arena-battle_counts-heatmap}
    \end{subfigure}
    \caption{Detailed human evaluation results from the arena study (Arena\textsubscript{\,Global})}
    \label{fig:arena}
\end{figure*}

Complete human evaluation results in terms of Bradley--Terry scores and final rankings for each arena dimension (content, language, and global quality) can be consulted in \cref{tab:arena-full}. Additionally, the detailed global win-rates and battle counts for each model pair are shown in \cref{fig:arena}. Note that the reported win-rates incorporate ties in the calculation, resulting in win and loss percentages that do not necessarily sum to 100\% for each model pair.
Note also that base models (i.e., \base{EN} and \base{EU}) were not included in the arena evaluation, as they are not capable of following instructions. However, their performance on benchmarks can be consulted in \cref{sap:benchmark-results}.

\begin{table*}[p]
    \centering
    \begin{tblr}{
    colspec={lrrlrlrlrlrl},
    cells={font=\footnotesize},
    row{1,2}={font=\footnotesize\bfseries},
    column{1,4,8}={rightsep=12pt},
    column{4,8,12}={l,font=\scriptsize,leftsep=2pt,preto=(,appto=)},
    column{6,10}={l,font=\scriptsize,leftsep=2pt},
    rowsep=0pt
}
    \toprule
     & \SetCell[c=3]{c} Arena\textsubscript{\,Global} &&& \SetCell[c=4]{c} Arena\textsubscript{\,Content} &&&& \SetCell[c=4]{c} Arena\textsubscript{\,Language} \\
     \cmidrule[r=-0.8]{2-4} \cmidrule[r=-0.8,l=-0.8]{5-8} \cmidrule[l=-0.8]{9-12}
     & Rank & \SetCell[c=2]{c} Bradley--Terry && \SetCell[c=2]{c} Rank && \SetCell[c=2]{c} Bradley--Terry && \SetCell[c=2]{c} Rank && \SetCell[c=2]{c} Bradley--Terry \\
    \midrule
    GPT-4o     
        & 1 & 1188 & +13/-17 & 1 & = & 1183 & +15/-13 & 1 & = & 1093 & +12/-10 \\
    3.5 Sonnet 
        & 2 & 1153 & +13/-21 & 2 & = & 1150 & +12/-17 & 3 & $\blacktriangledown$1 & 1082 & +11/-11 \\
    INS\textsubscript{\,EN,70B}\,C\textsubscript{\,EU}\,I\textsubscript{\,EN} 
        & 3 & 1141 & +15/-11 & 3 & = & 1127 & +10/-11 & 2 & $\blacktriangle$1 & 1083 & +13/-13 \\
    \SetRow{inst-yellow} INS\textsubscript{\,EN}\,C\textsubscript{\,EU}\,I\textsubscript{\,EN+EU} 
        & 5 & 1050 & +13/-14 & 4 & $\blacktriangle$1 & 1047 & +12/-12 & 4 & $\blacktriangle$1 & 1038 & +10/-8 \\
    \SetRow{inst-yellow} INS\textsubscript{\,EN}\,C\textsubscript{\,EU}\,I\textsubscript{\,EU} 
        & 5 & 1050 & +14/-11 & 5 & = & 1045 & +11/-13 & 6 & $\blacktriangledown$1 & 1034 & +8/-10 \\
    \SetRow{inst-yellow} INS\textsubscript{\,EN}\,C\textsubscript{\,EU}\,I\textsubscript{\,EN} 
        & 6 & 1038 & +13/-13 & 6 & = & 1031 & +15/-12 & 5 & $\blacktriangle$1 & 1036 & +11/-10 \\
    \SetRow{llama-blue} BAS\textsubscript{\,EN}\,C\textsubscript{\,EU}\,I\textsubscript{\,EN+EU} 
        & 7 & 1025 & +13/-11 & 7 & = & 1026 & +13/-13 & 10 & $\blacktriangledown$3 & 1019 & +7/-12 \\
    \SetRow{latxa-red} BAS\textsubscript{\,EU}\,C\textsubscript{\,EU}\,I\textsubscript{\,EN+EU} 
        & 8 & 1022 & +13/-8 & 8 & = & 1019 & +12/-11 & 9 & $\blacktriangledown$1 & 1021 & +10/-10 \\
    \SetRow{llama-blue} BAS\textsubscript{\,EN}\,C\textsubscript{\,EU}\,I\textsubscript{\,EN} 
        & 9 & 1017 & +12/-11 & 10 & $\blacktriangledown$1 & 1004 & +11/-14 & 7 & $\blacktriangle$2 & 1027 & +10/-9 \\
    \SetRow{latxa-red} BAS\textsubscript{\,EU}\,I\textsubscript{\,EN+EU} 
        & 10 & 1008 & +13/-13 & 9 & $\blacktriangle$1 & 1008 & +10/-13 & 12 & $\blacktriangledown$2 & 1008 & +9/-9 \\
    \SetRow{latxa-red} BAS\textsubscript{\,EU}\,C\textsubscript{\,EU}\,I\textsubscript{\,EN} 
        & 12 & 1005 & +17/-14 & 13 & $\blacktriangledown$1 & 989 & +16/-13 & 8 & $\blacktriangle$4 & 1026 & +10/-13 \\
    \SetRow{latxa-red} BAS\textsubscript{\,EU}\,I\textsubscript{\,EN} 
        & 12 & 1005 & +12/-14 & 11 & $\blacktriangle$1 & 1000 & +13/-13 & 11 & $\blacktriangle$1 & 1014 & +10/-10 \\
    \SetRow{latxa-red} BAS\textsubscript{\,EU}\,I\textsubscript{\,EU} 
        & 13 & 991 & +13/-16 & 12 & $\blacktriangle$1 & 990 & +13/-17 & 15 & $\blacktriangle$2 & 991 & +11/-10 \\
    \SetRow{llama-blue} BAS\textsubscript{\,EN}\,C\textsubscript{\,EU}\,I\textsubscript{\,EU} 
        & 14 & 985 & +12/-16 & 14 & = & 985 & +13/-15 & 13 & $\blacktriangle$1 & 1002 & +10/-10 \\
    \SetRow{latxa-red} BAS\textsubscript{\,EU}\,C\textsubscript{\,EU}\,I\textsubscript{\,EU} 
        & 15 & 982 & +12/-17 & 15 & = & 984 & +10/-14 & 14 & $\blacktriangle$1 & 994 & +10/-11 \\
    \SetRow{inst-yellow} INS\textsubscript{\,EN}\,C\textsubscript{\,EU} 
        & 16 & 942 & +9/-12 & 17 & $\blacktriangledown$1 & 935 & +8/-12 & 16 & = & 974 & +9/-8 \\
    \SetRow{inst-yellow} INS\textsubscript{\,EN}\,I\textsubscript{\,EN+EU} 
        & 17 & 936 & +14/-11 & 16 & $\blacktriangle$1 & 943 & +10/-13 & 18 & $\blacktriangledown$1 & 944 & +11/-10 \\
    \SetRow{inst-yellow} INS\textsubscript{\,EN}\,I\textsubscript{\,EU} 
        & 18 & 918 & +14/-14 & 18 & = & 924 & +11/-12 & 19 & $\blacktriangledown$1 & 943 & +12/-11 \\
    \SetRow{llama-blue} BAS\textsubscript{\,EN}\,I\textsubscript{\,EN+EU} 
        & 19 & 915 & +16/-14 & 19 & = & 922 & +15/-17 & 17 & $\blacktriangle$2 & 955 & +12/-10 \\
    \SetRow{llama-blue} BAS\textsubscript{\,EN}\,I\textsubscript{\,EU} 
        & 20 & 896 & +14/-16 & 20 & = & 909 & +12/-16 & 20 & = & 925 & +10/-13 \\
    \SetRow{llama-blue} \instruct{EN} 
        & 21 & 722 & +19/-17 & 21 & = & 766 & +17/-14 & 21 & = & 783 & +12/-12 \\
    \bottomrule
\end{tblr}

    \caption{Full results of human evaluation in the arena, across the three evaluation dimensions: content quality, linguistic quality and global preference. For convenience, 8B parameter-sized models are highlighted by backbone family (\colorbox{llama-blue}{\base{EN}}, \colorbox{latxa-red}{\base{EU}}, and \colorbox{inst-yellow}{\instruct{EN}}). Rankings for content and language include differences ({\scriptsize $\blacktriangle$/$\blacktriangledown$}) relative to the global ranking. Values in parenthesis indicate 90\% confidence intervals for Bradley--Terry scores.}
    \label{tab:arena-full}
\end{table*}

\end{document}